\documentclass{article} 
\usepackage{arxiv_diet,times}

\usepackage{amsmath,amsfonts,bm}

\def\eqref#1{equation~\ref{#1}}

\def\1{\bm{1}}

\DeclareMathAlphabet{\mathsfit}{\encodingdefault}{\sfdefault}{m}{sl}
\SetMathAlphabet{\mathsfit}{bold}{\encodingdefault}{\sfdefault}{bx}{n}

\DeclareMathOperator*{\argmin}{arg\,min}

\usepackage{booktabs}
\usepackage{amsmath}
\usepackage{amssymb}
\usepackage{makecell}
\usepackage{hyperref}
\hypersetup{hidelinks}
\usepackage{url}
\usepackage{graphicx}
\usepackage{wrapfig}
\usepackage{multirow}
\usepackage{textcomp}
\usepackage{algorithm}
\usepackage{algpseudocode}
\usepackage{subcaption}
\usepackage{enumitem}
\usepackage{placeins}
\usepackage{xcolor}
\usepackage{tikz}
\usetikzlibrary{arrows.meta,positioning}

\newcommand{\ours}{\textsc{DIET}}
\newcommand{\vbench}{VBench}
\newcommand{\sig}{\mathbf{s}}
\newcommand{\Kset}{\mathcal{K}}

\title{DIET: DeletIon-response Expert Trimming for Video Diffusion Transformers}

\author{Jiachang Zhang$^{1,2\ast}$, Teng Hu$^{2\ast}$, Bohao Feng$^{3}$\thanks{Equal Contribution} , Songhang Shen$^{2}$, Wenqiang Wang$^{4}$, \\ 
\textbf{Hongqian Deng$^{4}$, Ran Yi$^{2}$\thanks{Corresponding Author}}   \\
$^1$ Xi'an Jiao Tong University, China ~~~~~~
$^2$ Shanghai Jiao Tong University, China  \\
$^3$ Alibaba Token Hub, Alibaba Group, China ~~~~~~
$^4$ Alibaba Cloud Computing, China\\
\texttt{zjc010100000@stu.xjtu.edu.cn} ~~~
\texttt{\{hu-teng,1479014299,ranyi\}@sjtu.edu.cn} \\
\texttt{fengbohao.fbh@alibaba-inc.com} ~~~
\texttt{channing.wwq@alibaba-inc.com} \\
\texttt{hongqiandeng@foxmail.com}\\
}
\iclrfinalcopy 
\begin{document}

\maketitle

\begin{abstract}
Video diffusion transformers (DiTs) increasingly rely on mixture-of-experts (MoE) architectures, where only a sparse subset of a large expert bank is activated per token. While dynamic sparsity reduces active compute, it leaves the full expert storage footprint intact. Furthermore, conventional one-shot pruning criteria rely on static activation or routing statistics, failing to capture the layer-level re-routing behavior triggered after expert deletion. To make this post-deletion behavior observable without prohibitive cost, we record all expert outputs and router states across matched conditional and unconditional tokens in a single all-expert calibration pass. Consequently, replaying any single-expert deletion and its grouped re-routing reduces to tensor arithmetic on cached states, requiring zero additional model forward passes. Using these deletion responses, we formulate retained-expert selection as an Overall Diversity Loss (ODL): each pruned expert is matched with its nearest retained expert and the summed nearest-neighbor cosine distances define a set-level coverage loss under which deleted experts closer to retained ones cost less, directly prioritizing functional coverage. We solve this optimization in two stages: an intra-layer local search combining greedy initialization, single-swap refinement, and simulated annealing to select retained set under a fixed retention ratio, and an inter-layer regression-guided search that optimizes expert-count allocation across layers. Evaluated on LingBot-Video 30B-A3B, pruning $50\%$ of the experts ($6{,}144\!\to\!3{,}072$) without fine-tuning reduces the checkpoint footprint from $57$\,GB to $30$\,GB, enabling single-card deployment on a $48$\,GB GPU while generation quality remains essentially intact: we observe the official \vbench{} Total rise from $0.7941$ to $0.8115$ across a fixed $284$-case protocol. Across all other tested retention budgets, \ours{} consistently outperforms competitive baselines adapted from large language models. Code and project page are available at \url{https://zjc301.top/Video-DiT-MoE-Pruning-Page/}.
\end{abstract}

\section{Introduction}
\label{sec:intro}

Mixture-of-experts (MoE) layers route each token to a small subset of a large expert bank~\citep{jacobs1991adaptive,jordan1994hierarchical,shazeer2017moe}, and they have become a standard instrument for scaling: GShard, Switch and GLaM established that conditional computation adds capacity at a fraction of the dense compute~\citep{lepikhin2021gshard,fedus2022switch,du2022glam}, and many leading language models are MoEs today~\citep{jiang2024mixtral,dai2024deepseekmoe,deepseekv3,groeneveld2024olmoe}. Video generation has taken the same route: transformer backbones displaced U-Nets~\citep{peebles2023dit}, video diffusion scaled from short clips to systems such as Sora~\citep{ho2022video,blattmann2023svd,brooks2024sora}, and MoE now carries the leading video models: Seedance 2.0~\citep{seedance2.0} is the first video generation model to succeed through MoE scaling while remaining closed-source, while LingBot-Video~\citep{lingbot2026} is among the first open-source MoE video generation models. The DiT-MoE work had already shown that MoE layers scale a diffusion transformer to $16$B parameters in image generation~\citep{ditmoe2024}. But sparse activation saves only compute, leaving actual storage reduction a significant problem.

Removing whole experts is the most direct way to shrink that storage, because experts carry most of the parameters in the model. 
And the saving is proportional to how many experts go, so everything hinges on \emph{which} ones. Existing criteria typically determine retained sets via static activation statistics, geometric merging~\citep{ream2026}, or isolated output reconstruction: scalar scores rank experts by routing frequency or router-weighted activations~\citep{reap2026}, unified scoring families~\citep{hts2026,mone2025}, or coalitional coverage of retained sets~\citep{shape2026}. While more advanced methods address single-expert output reconstruction, they remain fundamentally limited: they either assume layer damage can be ranked independently, or optimize for point-to-point activation reconstruction, which our analysis shows to be inferior to preserving response-space diversity. Furthermore, exhaustive evaluation of multi-expert deletions is prohibitively expensive at this scale. 

\begin{figure}[t]
\centering
\includegraphics[width=\textwidth]{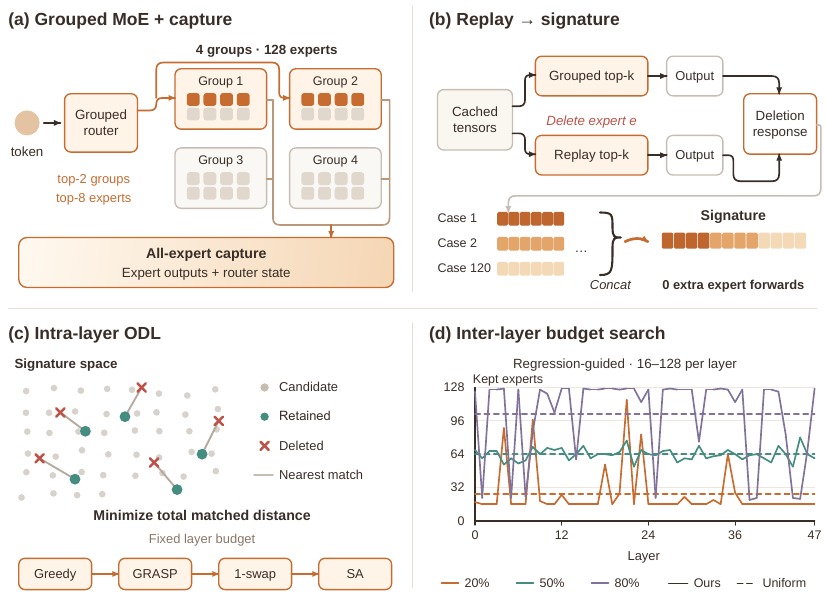}
\vspace{-0.10in}
\caption{(a) Grouped routing in LingBot-Video: each token activates eight of the $128$ experts of a layer through grouped top-$2$ routing, and a single all-expert calibration records every expert output and the router state. (b) Deleting an expert and replaying the grouped top-$k$ reproduces the deletion at the captured states from cached tensors, without extra expert forwards, and the case-level responses concatenate into a \emph{deletion-response signature}. (c) The overall diversity loss (ODL) matches every deleted expert to its nearest retained expert in signature space and sums the matched distances, and the retained set is optimized under a given per-layer budget. (d) The regression-guided search allocates the retention ratio across layers: solid curves, our allocations at $20\%$, $50\%$ and $80\%$ retention; dashed, uniform references.}
\vspace{-0.16in}
\label{fig:teaser}
\end{figure}

These observations naturally lead to two central questions: 
\emph{First, can post-deletion layer dynamics be observed without incurring prohibitive computational costs from repeated forward passes?}
\emph{Second, on a large-scale video DiT-MoE where fine-tuning is computationally inaccessible, can we formulate expert retention directly from these measured dynamics to preserve generation quality?}
Resolving these challenges requires making the fine-grained re-routing behavior observable at a low cost, and designing an objective that respects the non-additive geometry of diffusion representations rather than relying on isolated scalar heuristics (Figure~\ref{fig:teaser}).

To address these challenges, we introduce \ours{}, a training-free framework that converts observable post-deletion responses into principled expert retention decisions for large-scale video MoEs. Rather than estimating expert importance from static weights, activations, or routing frequencies, \ours{} directly characterizes how the layer behaves after each expert is removed and re-routing takes place. Specifically, \textbf{we first develop a frozen-state replay mechanism} that records all expert outputs and router states in a single instrumented calibration pass, enabling all $6{,}144$ layer-local deletions to be reconstructed through tensor arithmetic with zero additional model forward passes. These counterfactual responses are pooled into expert-level deletion-response signatures, which jointly capture expert functionality and its interaction with layer-local re-routing. \textbf{We then formulate intra-layer pruning as directional functional coverage} and introduce an Overall Diversity Loss (ODL) that selects retained experts by preserving the response directions induced by deleted ones through efficient local combinatorial search. Beyond individual layers, \textbf{we introduce a regression-guided inter-layer budget search} that learns the relationship between layer-wise retention configurations and validation quality, allowing the pruning budget to adapt to non-uniform layer sensitivities without exhaustively evaluating candidate models. Together, these components enable \ours{} to prune $50\%$ of the experts in LingBot-Video 30B-A3B, reducing the checkpoint footprint from $57$,GB to $30$,GB while improving the official \vbench{} Total from $0.7941$ to $0.8115$ and consistently outperforming ported LLM pruning baselines across all tested budgets.

Our contributions are summarized as follows. \emph{\textbf{1)}} We propose \ours{}, a training-free expert pruning framework that directly uses measured post-deletion and re-routing responses to formulate expert retention as directional functional coverage. \emph{\textbf{2)}} We develop a frozen-state replay mechanism that evaluates all $6{,}144$ layer-local deletions with zero additional model forward passes, together with an ODL objective for efficient intra-layer retained-set selection. \emph{\textbf{3)}} We introduce a regression-guided inter-layer budget search that captures non-uniform layer sensitivities and identifies tailored retention allocations, yielding a $1.2$-point gain over uniform budgets. \emph{\textbf{4)}} We validate \ours{} on LingBot-Video 30B-A3B, where $50\%$ expert pruning reduces the checkpoint footprint from $57$,GB to $30$,GB while increasing the official \vbench{} Total from $0.7941$ to $0.8115$ and outperforming ported LLM pruning baselines across all tested budgets.
\section{Related Work}
\label{sec:related}

\subsection{Mixture-of-Experts: From Language to Video Diffusion}
Sparse computation through mixture-of-experts (MoE) provides an established path to decouple parameter-based model capacity from inference compute~\citep{jacobs1991adaptive,jordan1994hierarchical,shazeer2017moe,lepikhin2021gshard,fedus2022switch}. In natural language processing, sparsely gated MoE architectures underpin leading frontier models~\citep{du2022glam,dai2024deepseekmoe,jiang2024mixtral,deepseekv3,groeneveld2024olmoe}, where scaling dynamics have been thoroughly characterized~\citep{clark2022unified}. To suppress communication overhead in massive distributed deployments, modern variants frequently enforce group-limited routing, confining token dispatching to a restricted subset of expert groups~\citep{lepikhin2021gshard,deepseekv3}.

Recently, this conditional scaling paradigm has crossed into video generation, transitioning from U-Net video diffusion~\citep{ho2022video,blattmann2023svd} to diffusion transformers (DiTs)~\citep{peebles2023dit,brooks2024sora} toward sparse backbones. Seedance~2.0 pioneered the industrial deployment of MoE architectures for scalable video synthesis~\citep{seedance2.0}, while LingBot-Video~\citep{lingbot2026} and MAGI-2~\citep{magi2} established the first open-source video DiT-MoE backbones utilizing grouped and ultra-fine-grained routing, respectively. However, dynamic routing sparsity merely reduces active FLOPs per step; the entire expert bank must permanently reside in device memory. 
In multi-billion-parameter video DiTs, this parameter footprint poses an acute deployment barrier, requiring multi-GPU tensor-parallel or FSDP sharding simply to hold quiescent weights.

\subsection{MoE Pruning and Retained Expert Selection}
Pruning techniques compress over-parameterized networks by eliminating redundant structures~\citep{lecun1990optimal,hassibi1993second,han2015learning,frankle2019lottery,moepruner2024}. Structured pruning compresses networks by eliminating regular parameter blocks to achieve native hardware savings~\citep{ma2023llmpruner}. While early methods focus on channels or attention heads~\citep{li2016pruning,xia2024sheared}, expert pruning acts as a macro-structural compression paradigm in MoEs, operating at the granularity of entire sub-networks to achieve physical memory savings strictly proportional to the excised expert count. Prior work on MoE compression falls broadly into three paradigms:

\textbf{Static scalar scoring.} 
Early criteria rank experts via aggregated statistics, such as historical routing frequency, router-weighted activation norms~\citep{reap2026}, or frequency--variance interactions~\citep{mone2025}; more recent formulations unify such signals, scoring experts through routing frequency, gate weighting, and activation strength~\citep{hts2026}. While computationally trivial, static scalar metrics fundamentally assume expert utility to be context-independent, failing to account for the dynamic re-routing triggered when an expert is removed.

\textbf{Re-evaluation and reconstruction.} 
To capture deletion impacts, advanced approaches assess post-removal layer states: NAEE~\citep{lu-etal-2024-experts} caches layer input-output token pairs to search for subsets that minimize replayed output discrepancy. We port its reconstruction criterion onto our cached activations as a strong fidelity baseline (Section~\ref{sec:comparison}). However, these methods are primarily designed for language MoEs without grouped routing. On fine-grained video DiTs, combinatorial replay is computationally prohibitive at this scale; more fundamentally, enforcing strict activation reconstruction serves as a suboptimal surrogate that overlooks directional manifold coverage in generative diffusion.

\textbf{Geometric merging and interaction modeling.} 
Other work exploits representation geometry or inter-expert dependencies. Expert merging fuses functionally redundant weights~\citep{ream2026}, though recent findings suggest that geometric proximity in weight or output space does not necessarily imply contextual interchangeability~\citep{coherentoverlap2026}. Other strategies model inter-expert dynamics via second-order Taylor approximations~\citep{hope2026}, dynamic inference-time token re-routing~\citep{sere2026}, game-theoretic Shapley values~\citep{shapleymoe2025,shape2026}, coverage-aware selection across calibration sources~\citep{tbcoverage2026}, or trajectory-guided global pruning~\citep{pathfinder2025}. 

\textbf{The gap in Video DiT-MoEs.} 
Existing MoE pruning strategies are developed primarily for autoregressive language models, optimizing for teacher-forced perplexity, token-level cross-entropy, or intermediate activation reconstruction. These formulations do not transfer cleanly to diffusion transformers, where multi-step denoising trajectories, classifier-free guidance (CFG), and non-additive post-deletion response manifolds complicate classical reconstruction objectives. Furthermore, prior interaction models commonly approximate dependencies only to pairwise order. In contrast, \ours{} captures the exact grouped re-routing at the frozen calibration state, up to numerical precision, induced by each expert deletion, with zero extra forward passes, framing combinatorial retained set selection as holistic geometric coverage over these measured deletion signatures.
\section{Methodology}
\label{sec:method}

Expert pruning of a released video DiT-MoE over a given retention ratio naturally decomposes into two distinct yet interdependent sub-problems: \emph{which} experts each layer keeps, and \emph{how many} experts each layer may keep. The first requires that the bank contains redundancy. Our analyses on the routing side show its potential compressibility: routing counts are concentrated, with a long tail of rarely used experts (Appendix~\ref{app:routing}). However, raw routing frequencies alone cannot determine which experts to prune, because a deletion causes re-routing: it changes the grouped top-$k$ decision, the tokens the deleted expert served move to other experts, and its gate weight is redistributed over the retained experts (Appendix~\ref{app:model}). Any criterion that scores experts independently of this re-routing fails to capture the underlying layer dynamics, and the statistics that such criteria rely on may correlate broadly in aggregate, but remain unreliable at individual layers (Section~\ref{sec:ablation}). Our solution is to measure the re-routing itself. Section~\ref{sec:signature} turns one instrumented all-expert capture into a measurable deletion response for every expert and pools those responses into signatures; Section~\ref{sec:objective} translates the signatures into a per-layer selection rule and solves it; Section~\ref{sec:budget} chooses how many experts each layer keeps, where the two decisions interact.

\subsection{Deletion-Response Signatures via Frozen State Replay}
\label{sec:signature}

Simulating leave-one-out expert removals is prohibitively expensive if done naively: across $L=48$ layers and $E=128$ experts per layer, brute-force simulation requires at least $6{,}144$ forward passes. We bypass this computational barrier through an \emph{all-expert calibration pass} followed by \emph{layer-local frozen-state replay}. 

Specifically, we run a single instrumented calibration sweep over the $120$ cases of the calibration set (Appendix~\ref{app:protocol}), recording router logits, gate weight, routing selections, and full expert output tables across the sampled tokens of every layer. Although this calibration evaluates all $128$ experts per token (incurring $\approx 16\times$ the expert compute of a normal pass), the cached state for replay is $\approx 375$\,GB in half precision (\texttt{fp16} expert outputs with \texttt{bf16} gates), paid strictly once. Subsequently, simulating the removal of any expert and recomputing the grouped top-$k$ routing reduces entirely to tensor arithmetic over cached states, requiring zero additional model forward passes up to numerical precision (Appendix~\ref{app:injection}). A summary of the complete pipeline's compute budget is given in Appendix~\ref{app:compute}.

\paragraph{Deletion-response signature.}
For a given token $x$ at layer $\ell$, we define the \emph{deletion response} of expert $e$ as the perturbation in layer output induced by its removal:
\begin{equation}
    \Delta_{\ell,e}(x) \;=\; y_{[E]\setminus\{e\}}(x) - y_{[E]}(x),
    \label{eq:response}
\end{equation}
where $y_{\mathcal{S}}(x)$ denotes the layer output restricted to the active expert bank $\mathcal{S}$. Crucially, $\Delta_{\ell,e}(x)$ inherently accounts for the discrete token re-routing and gate re-normalization triggered by the deletion (Appendix~\ref{app:model}). For classifier-free guidance (CFG), $\Delta_{\ell,e}(x)$ is computed directly on the guidance-weighted outputs.

To construct a fixed-dimensional functional representation for each expert, we average the perturbation across tokens within each calibration case and concatenate the resulting case-wise vectors into a \emph{deletion-response signature}:
\begin{equation}
    \sig_{\ell,e} \;=\; \operatorname*{Concat}_{c=1}^{C}\left[\frac{1}{|\mathcal{T}_c|}\sum_{t\in\mathcal{T}_c}\Delta_{\ell,e}(x_{c,t})\right],
    \label{eq:signature}
\end{equation}
where $\mathcal{T}_c$ denotes the set of sampled token positions for case $c$. 

Two geometric properties of $\sig_{\ell,e}$ govern our downstream formulation. First, because each leave-one-out counterfactual is measured with all complementary experts present at the captured state, the signature isolates the marginal utility of expert $e$ conditioned on its contextual interactions within the layer, rather than evaluating its isolated behavior. Second, the signature is utilized strictly for its directional orientation rather than its Euclidean magnitude: because concurrent expert removals violate linear superposition (Section~\ref{sec:ablation}), optimizing for absolute perturbation size fails to reliably reflect joint post-pruning dynamics. Instead, we use directional alignment as a surrogate for functional interchangeability.

\subsection{Selecting Retained Experts by Overall Diversity Loss}
\label{sec:objective}

Retained set selection should be guided by deletion-response signatures rather than conventional magnitude proxies. Minimizing the norm of summed perturbations or ranking experts by activation energy substantially degrades semantic performance (Section~\ref{sec:ablation}), as single-deletion responses violate linear superposition. Moreover, the signature space admits no informative low-rank projection that classical compression could exploit; its exploitable structure is instead a sparse tail of functionally aligned expert pairs (Appendix~\ref{app:corr}). Pruning is therefore formulated as selecting \emph{which} directions to eliminate: each pruned expert is matched with its nearest surviving counterpart, as near as the surviving set permits, so that experts whose functional perturbations are already covered by retained experts can be removed at little cost.

To this end, we introduce the Overall Diversity Loss (ODL), which pairs each pruned expert with its nearest surviving counterpart in signature space and penalizes functional divergence via cosine distance:
\begin{equation}
    \Kset_\ell^\star \;=\; \argmin_{\substack{\Kset_\ell \subseteq [E] \\ |\Kset_\ell| = k_\ell}}
    \sum_{e \notin \Kset_\ell} \min_{f \in \Kset_\ell} d\big(\sig_{\ell,e}, \sig_{\ell,f}\big),
    \qquad
    d(u,v) = 1 - \frac{\langle u, v\rangle}{\|u\|\,\|v\|}.
    \label{eq:odl}
\end{equation}
Equation~\ref{eq:odl} formulates an assignment-cost ($k$-median) problem optimizing functional coverage. Crucially, the cosine metric is scale-invariant, assessing directional redundancy rather than raw activation magnitude. Furthermore, summing individual matched distances prevents artificial cancellation between opposing directional responses, avoiding a primary pitfall of global norm-based objectives (Section~\ref{sec:ablation}).

\paragraph{Combinatorial optimization.}
We solve Equation~\ref{eq:odl} layer by layer across candidate retention budgets. The solver initializes via a feasible greedy selection with randomized restarts (GRASP), followed by single-swap local search and simulated annealing refinement. This combinatorial search yields substantial objective improvements over naive greedy selection (Table~\ref{tab:ablation}). Precomputing these optimized retained sets across candidate budgets yields an empirical Pareto frontier for each layer, serving as the lookup foundation for inter-layer allocation (Section~\ref{sec:budget}).

\subsection{Inter-Layer Budget Allocation}
\label{sec:budget}

Because individual layers exhibit non-uniform sensitivity to expert pruning, per-layer budgets $\{k_\ell\}$ should be allocated globally rather than uniformly. Given the precomputed layer-wise ODL frontiers, any layer retention vector $\mathbf{k} = [k_1, \dots, k_L]$ can be mapped directly to concrete retained sets via constant-time lookups (Appendix~\ref{app:budget}).

We formulate inter-layer budget allocation through a fitting-and-proposal loop. Near the unpruned operating point, we assume that the benchmark response varies approximately linearly with the per-layer retention counts. For a candidate allocation $\mathbf{k} = [k_1, \dots, k_L]$, each dimension score is modeled as
\begin{equation}
    y_d \;=\; \mathbf{w}_d^\top \mathbf{k} + b_d,
    \label{eq:surrogate}
\end{equation}
where $y_d$ is the measured score of dimension $d$ and $k_\ell$ is the number of experts retained at layer $\ell$. The coefficients are fitted by a regularized linear surrogate ($\lambda = 1$) over the accumulated end-to-end evaluation corpus with an unpenalized intercept, using the uncentered design matrix, $\mathbf{w}_d = (\mathbf{K}^\top\mathbf{K} + \lambda\mathbf{I})^{-1}\mathbf{K}^\top(\mathbf{y}_d - \bar{y}_d\mathbf{1})$ and $b_d = \bar{y}_d - \bar{\mathbf{k}}^\top\mathbf{w}_d$. The fitted surrogate then proposes candidate allocations by solving a constrained linear program: it maximizes the predicted Total subject to the global budget $\sum_\ell k_\ell = B$ ($B=3{,}072$ at $50\%$), per-layer feasibility bounds, a one-sided per-dimension degradation floor that bounds predicted regressions against the unpruned profile, and a Total-preservation constraint enforced as a hard floor at retention budgets of $50\%$ and above; the continuous optimum is then integerized by a small mixed-integer step. The cross-layer associations this fitted analysis exposes, and their end-to-end payoff, are reported in Section~\ref{sec:analysis} (Appendix~\ref{app:budgetdetails}).

Each proposed budget allocation is subsequently validated via full end-to-end video generation and official benchmark evaluation, with candidate trajectories incrementally expanding the surrogate set. All reported results reflect empirical end-to-end evaluations rather than surrogate predictions. Finally, the chosen retained mask is materialized into a physically compacted checkpoint, eliminating pruned parameters and reducing weight residency without altering routing logic (Appendix~\ref{app:injection}).
\section{Experiments}
\label{sec:experiments}

\subsection{Main Results}
\label{sec:main}

\paragraph{Experimental setup.}
\label{sec:setup}
We evaluate our framework on the open-source LingBot-Video 30B-A3B diffusion transformer ($48$ layers, $128$ experts per layer, grouped top-$8$ routing across four groups, detailed in Appendix~\ref{app:model}). We prune $50\%$ of the expert bank ($3{,}072$ retained experts out of $6{,}144$). Videos are synthesized at $240\times416$ resolution and evaluated using the official \vbench{} scorer~\citep{huang2024vbench} across a fixed set of $284$ prompts with matched random seeds to ensure paired comparisons (Appendix~\ref{app:protocol}). Crucially, because expert removal triggers discrete token re-routing, the pruned model generates plausible alternative visual motions rather than strictly reproducing baseline trajectories. Consequently, we assess performance via perceptual and semantic benchmark quality rather than point-wise reconstruction metrics (e.g., PSNR). All evaluated methods share identical $120$-case calibration data, and the surrogate budget search is calibrated on a $142$-case corpus drawn from the same prompt distribution (Appendix~\ref{app:budgetdetails}).

\paragraph{Primary results.}
Table~\ref{tab:ours} summarizes the main performance envelope. At $50\%$ expert retention without any retraining, \ours{} raises the official \vbench{} Total from $0.7941$ to $0.8115$, with both visual Quality and Semantic scores outperforming the dense baseline. One possible explanation for this counter-intuitive improvement is the implicit regularization induced by structured sparsification. Importantly, the positive trend transfers to the held-out partition, although the held-out-only interval is not individually significant at 95\% ($+0.0142$, Appendix~\ref{app:budgetdetails}). 

In terms of deployment efficiency, active FLOPs remain unchanged by construction, while the physical checkpoint footprint is halved from $57$\,GB to $30$\,GB. This reduction enables a model replica to reside entirely within a single $48$\,GB workstation GPU, eliminating multi-GPU sharding overheads (Table~\ref{tab:compression}, Appendix~\ref{app:compression}). The pruned checkpoint remains robust when scaled to native $480$p resolution, showing no systematic qualitative artifacts (Appendices~\ref{app:resolution} and~\ref{app:qual}).

\begin{table}[tb]
\centering
\caption{Performance across retention budgets on official \vbench{} ($284$-case paired protocol). Expert budgets correspond to $1{,}229$, $3{,}072$, and $4{,}915$ retained experts out of $6{,}144$.}
\label{tab:ours}
\small
\setlength{\tabcolsep}{8pt}
\begin{tabular}{lrrr}
\toprule
Retention Budget & Total $\uparrow$ & Quality $\uparrow$ & Semantic $\uparrow$ \\
\midrule
Unpruned (100\%) & 0.7941 & 0.8125 & 0.7204 \\
\midrule
$20\%$ ($1{,}229$ experts) & 0.7596 & 0.7958 & 0.6145 \\
$50\%$ ($3{,}072$ experts) & \textbf{0.8115} & 0.8325 & \textbf{0.7278} \\
$80\%$ ($4{,}915$ experts) & 0.8061 & \textbf{0.8331} & 0.6980 \\
\bottomrule
\end{tabular}
\end{table}

\begin{figure}[tb]
\centering
\includegraphics[width=1.0\textwidth]{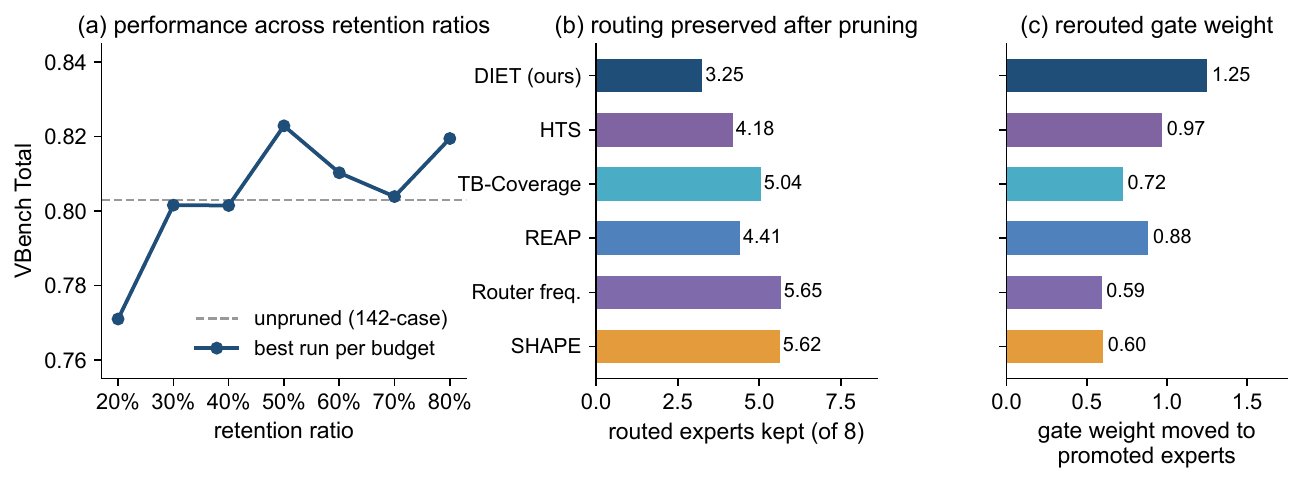}
\vspace{-0.12in}
\caption{Primary evaluation and routing dynamics. (a) \vbench{} Total across candidate retention budgets on the $142$-case budget-search set (dashed line denotes the dense baseline). (b) Number of original routed experts retained per token (out of 8). (c) Reallocated gating weight assigned to promoted surviving experts (averaged over $120$ calibration cases, $48$ layers). Counter-intuitively, \ours{} induces the highest routing perturbation while achieving the highest benchmark scores, demonstrating that routing preservation is a suboptimal objective.}
\vspace{-0.16in}
\label{fig:results}
\end{figure}

\subsection{Comparison Experiment}
\label{sec:comparison}

We compare \ours{} against established one-shot MoE pruning paradigms ported from large language models, including static router-statistic rankings (Routing Frequency, REAP), unified scoring (HTS), coalition-based selection (SHAPE), cross-register coverage (TB-Coverage), and replayed-reconstruction selection (NAEE-Style Recon), which optimizes the deletion-fidelity objective itself through the same solver used for \ours{}. All baselines are evaluated under identical calibration states, budgets, and scoring protocols across both searched and uniform layer allocations (Table~\ref{tab:main}). This design separates two readings. The uniform block is a criterion-level matched comparison: it isolates each retained-set criterion under one identical per-layer allocation. The searched block instead measures the end-to-end payoff of the complete \ours{} pipeline, criterion plus regression-guided allocation. The gaps observed there therefore describe the full pipeline, and are not a claim that the ported criteria would remain behind by the same margins under their own tuned allocations.

Across all retention regimes, \ours{} establishes the Pareto-optimal frontier. At $50\%$ retention, the strongest ported baseline is HTS, whose strongest variant attains a Total of $0.7921$, still lagging behind \ours{} ($0.8115$) by $1.94$ points. DIET also remains the strongest method under uniform layer allocations. NAEE-Style Recon is the reconstruction-criterion port: with the identical capture, budgets and solver, it reduces the replayed output error by about $40\%$ relative to \ours{} (Appendix~\ref{app:baselines}), yet its benchmark Total remains $2.80$ points below \ours{}.

\begin{table}[tb]
\centering
\caption{Official \vbench{} comparative evaluation ($284$-case protocol, identical sampling seeds and evaluation pipeline). Retention budgets correspond to $1{,}229$ ($20\%$), $3{,}072$ ($50\%$), and $4{,}915$ ($80\%$) surviving experts.}
\label{tab:main}
\small
\setlength{\tabcolsep}{5pt}
\begin{tabular}{lccccccccc}
\toprule
& \multicolumn{3}{c}{$20\%$ Retention} & \multicolumn{3}{c}{$50\%$ Retention} & \multicolumn{3}{c}{$80\%$ Retention} \\
\cmidrule(lr){2-4}\cmidrule(lr){5-7}\cmidrule(lr){8-10}
Method & Total & Qual. & Sem. & Total & Qual. & Sem. & Total & Qual. & Sem. \\
\midrule
\multicolumn{10}{@{}l}{\emph{Uniform Layer Budget}} \\
\ours{} (Ours) & \textbf{0.7370} & 0.7742 & \textbf{0.5883} & \textbf{0.7996} & 0.8184 & \textbf{0.7243} & \textbf{0.8052} & \textbf{0.8311} & 0.7013 \\
REAP & 0.6756 & 0.7645 & 0.3204 & 0.7807 & 0.8033 & 0.6901 & 0.7943 & 0.8171 & 0.7028 \\
SHAPE & 0.6647 & 0.7507 & 0.3209 & 0.7374 & 0.7662 & 0.6223 & 0.7850 & 0.8081 & 0.6929 \\
HTS & 0.6909 & 0.7801 & 0.3342 & 0.7921 & \textbf{0.8252} & 0.6595 & 0.8025 & 0.8274 & 0.7027 \\
TB-Coverage & 0.6815 & 0.7703 & 0.3265 & 0.7650 & 0.7976 & 0.6344 & 0.7982 & 0.8235 & 0.6970 \\
Router Frequency & 0.6406 & 0.7363 & 0.2580 & 0.7437 & 0.7802 & 0.5978 & 0.7921 & 0.8186 & 0.6859 \\
NAEE-Style Recon & 0.7007 & \textbf{0.7829} & 0.3720 & 0.7810 & 0.8147 & 0.6461 & 0.8028 & 0.8230 & \textbf{0.7217} \\
\midrule
\multicolumn{10}{@{}l}{\emph{Regression-Guided Layer Budget}} \\
\ours{} (Ours) & \textbf{0.7596} & \textbf{0.7958} & \textbf{0.6145} & \textbf{0.8115} & \textbf{0.8325} & \textbf{0.7278} & \textbf{0.8061} & \textbf{0.8331} & 0.6980 \\
REAP & 0.6931 & 0.7625 & 0.4159 & 0.7761 & 0.8081 & 0.6481 & 0.8020 & 0.8289 & 0.6941 \\
SHAPE & 0.6693 & 0.7469 & 0.3589 & 0.7471 & 0.7792 & 0.6187 & 0.7942 & 0.8166 & 0.7047 \\
HTS & 0.7216 & 0.7844 & 0.4704 & 0.7921 & 0.8234 & 0.6668 & 0.8019 & 0.8290 & 0.6937 \\
TB-Coverage & 0.7202 & 0.7901 & 0.4405 & 0.7711 & 0.8027 & 0.6448 & 0.8038 & 0.8273 & \textbf{0.7099} \\
Router Frequency & 0.6725 & 0.7534 & 0.3489 & 0.7485 & 0.7804 & 0.6207 & 0.7954 & 0.8182 & 0.7042 \\
NAEE-Style Recon & 0.7112 & 0.7725 & 0.4662 & 0.7835 & 0.8147 & 0.6589 & 0.8022 & 0.8267 & 0.7041 \\
\bottomrule
\end{tabular}
\end{table}

\subsection{Analyses}
\label{sec:analysis}

\paragraph{Routing sparsity versus functional redundancy.}
Routing statistics reveal that expert utilization is skewed but lacks dead capacity. Across $120$ calibration runs, the layer-wise Gini coefficient averages $0.346$ (range $0.100$--$0.576$), with the top $10\%$ of experts routing $23.1\%$ of tokens and normalized routing entropy remaining high ($0.955$, Appendix~\ref{app:routing}). The expert bank thus forms a continuous long tail rather than a partition of inactive units. Consequently, frequency-based criteria can prune moderately active, functionally critical experts, conflating routing volume with post-removal layer resilience.

\begin{wrapfigure}[13]{r}{0.50\textwidth}
\vspace{-0.08in}
\centering
\includegraphics[width=0.48\textwidth]{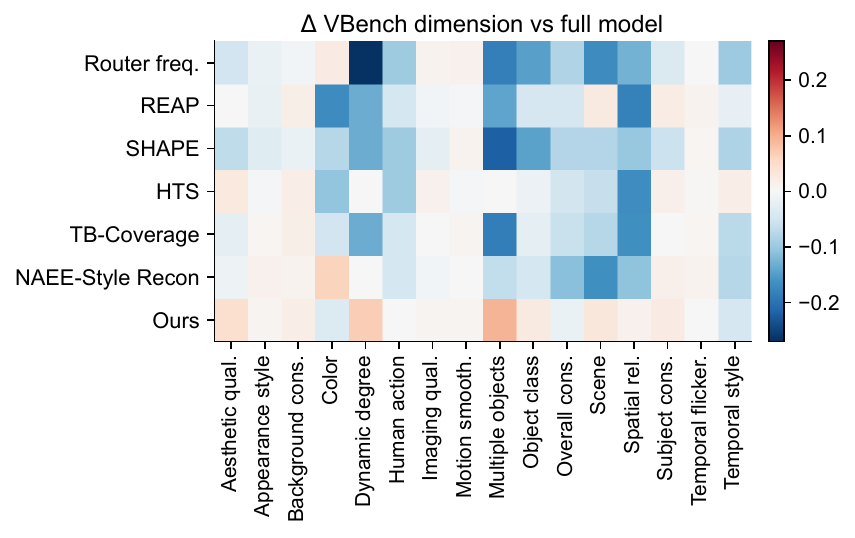}
\par\vspace{2pt}
{\footnotesize\captionof{figure}{Per-dimension deviation from the unpruned baseline at $50\%$ retention.\label{fig:delta}}}
\end{wrapfigure}

\paragraph{Rethinking routing disruption and reconstruction objectives.}
Crucially, empirical results challenge the conventional intuition that pruning should minimize routing disturbance. As illustrated in Figure~\ref{fig:results}(b--c), we observe \ours{} produces the most substantial routing perturbation among all evaluated masks---retaining only $3.25$ of the $8$ originally selected experts per token and reallocating $1.25$ units of gating weight. Remarkably, the best-performing mask also induces the largest routing disruption. This helps explain why strict reconstruction can be limiting in video MoEs. Enforcing point-wise layer output preservation (the analog of PSNR minimization) misaligns with the generative diffusion mechanism: an expert deletion triggers discrete token dispatch to alternative experts, modulating subsequent denoising steps rather than corrupting a deterministic signal. Furthermore, because inter-expert interactions are non-linear, local reconstruction errors do not compose linearly across cascaded transformer blocks. Instead of attempting to reproduce original activations, DIET preserves directional coverage, which deletion-response signatures directly capture. As visualized in Figure~\ref{fig:delta}, this preserves fine-grained semantic integrity where classical baselines suffer severe degradation.

\paragraph{Cross-layer budget structure.} Layer-wise retention allocations exhibit structured associations with specific visual dimensions (Figure~\ref{fig:layerdim}): expert retention in early blocks correlates with spatial coherence and scene composition, middle blocks with dynamic degree and appearance style, and deep blocks with color fidelity and temporal subject consistency. Because fixed global budgets impose cross-layer capacity trade-offs, this functional heterogeneity is precisely what non-uniform allocation exploits, contributing the $1.2$-point end-to-end gain over uniform allocation reported in Table~\ref{tab:ablation}.

\subsection{Ablation Studies}
\label{sec:ablation}

We isolate the contribution of each algorithmic component under the $50\%$ retention budget (Table~\ref{tab:ablation}).

\begin{wrapfigure}[15]{r}{0.46\textwidth}
\vspace{-0.15in}
\centering
\resizebox{\linewidth}{!}{%
\setlength{\tabcolsep}{3.5pt}
\begin{tabular}{llr}
\toprule
Component & Configuration & Total $\uparrow$ \\
\midrule
\multirow{3}{*}{\emph{Space}} 
& Router Frequency & 0.7485 \\
& Router Score & 0.7813 \\
& Output Space & 0.7883 \\
\midrule
\multirow{3}{*}{\emph{Objective}}
& Summed-Response Norm & 0.7571 \\
& PCA Energy ($r{=}32$) & 0.7849 \\
& D-Optimal Volume & 0.8007 \\
\midrule
\multirow{2}{*}{\emph{Optimization}}
& Uniform Layer Budget & 0.7996 \\
& Greedy Init Only & 0.7896 \\
\midrule
\textbf{Proposed} & \textbf{\ours{} (Full)} & \textbf{0.8115} \\
\bottomrule
\end{tabular}%
}
\par\vspace{2pt}
{\footnotesize\captionof{table}{Ablation analysis on representation spaces, objectives, and optimization strategies at $50\%$ retention.\label{tab:ablation}}}
\end{wrapfigure}

\paragraph{Representation space.}
Evaluating experts outside of their layer-level operational context leads to steep performance drops. Substituting deletion signatures with isolated expert outputs or router affinities degrades the Total score by $2.32$--$3.02$ points, while relying on static routing frequency induces a $6.30$-point drop. Isolated output similarity~\citep{coherentoverlap2026} ignores the contextual interactions and re-routing mechanisms that determine post-pruning layer behavior. Appendix~Fig.~\ref{fig:galleryadd} illustrates the frame-level degradation.

\paragraph{Selection objective.}
Minimizing the aggregate norm of single-deletion responses (an additive formulation) produces the steepest drop ($-5.44$ points), collapsing spatial coherence dimensions from $0.63$ to $0.33$. This failure stems directly from the breakdown of linear superposition: as shown in Appendix~\ref{app:ablation}, the cosine similarity between a joint multi-expert deletion response and the sum of constituent single-deletion responses drops from $1.00$ at $k{=}1$ to $0.66$ at our operating point, yielding an error of $0.95$. Similarly, low-rank subspace approximations fall short: D-optimal volume selection and PCA energy retention trail \ours{} by $1.08$ and $2.66$ points, respectively. Consistent with the observed flat singular spectrum, dimensionality reduction discards vital sparse tail directions that ODL explicitly preserves.

\paragraph{Optimization and allocation.}
Finally, both algorithmic stages provide measurable gains. Greedy initialization leaves isolated tail directions uncovered; refining the partition via single-swap local search and simulated annealing recovers these modes, substantially lowering the combinatorial objective. At the inter-layer level, substituting the regression-guided allocation with a uniform budget costs $1.19$ points, confirming that layer-wise tolerance to expert removal is non-uniform and can be effectively exploited via surrogate optimization.
\section{Conclusion and Limitations}
\label{sec:conclusion}

Our work demonstrates that the expert bank of a deployed video diffusion transformer can be compressed by half without post-pruning fine-tuning or weight updates. Frozen-state replay via an instrumented all-expert calibration sweep makes combinatorial leave-one-out deletion studies tractable with no extra model forwards. Because joint post-deletion responses violate linear superposition and induce discrete token re-routing, retained set selection maximizes directional functional coverage instead of strict activation reconstruction. Coupling intra-layer Overall Diversity Loss (ODL) with regression-guided inter-layer budget allocation enables LingBot-Video 30B to operate on a single $48$\,GB GPU, halving the checkpoint footprint while generation quality remains essentially intact, with the official \vbench{} Total observed to rise ($0.8115$ vs.\ $0.7941$).

\paragraph{Limitations.}
First, our layer-local ODL decomposition does not track cascading activation drift across downstream blocks. In addition, our findings rest on one representative open-source DiT-MoE backbone; cross-architecture generalization remains open. Finally, the front-loaded calibration and search compute scales down substantially for a single retention point (Appendices~\ref{app:calibsize} and~\ref{app:budget}).

\bibliography{arxiv_diet}

@inproceedings{shazeer2017moe,
  title={Outrageously Large Neural Networks: The Sparsely-Gated Mixture-of-Experts Layer},
  author={Shazeer, Noam and Mirhoseini, Azalia and Maziarz, Krzysztof and Davis, Andy and Le, Quoc and Hinton, Geoffrey and Dean, Jeff},
  booktitle={International Conference on Learning Representations},
  year={2017},
}

@article{fedus2022switch,
  title={Switch Transformers: Scaling to Trillion Parameter Models with Simple and Efficient Sparsity},
  author={Fedus, William and Zoph, Barret and Shazeer, Noam},
  journal={Journal of Machine Learning Research},
  volume={23},
  number={120},
  pages={1--39},
  year={2022},
}

@inproceedings{huang2024vbench,
  title={{VBench}: Comprehensive Benchmark Suite for Video Generative Models},
  author={Huang, Ziqi and He, Yinan and Yu, Jiashuo and Zhang, Fan and Si, Chenyang and Jiang, Yuming and Zhang, Yuanhan and Wu, Tianxing and Jin, Qingyang and Chanpaisit, Nattapol and Wang, Yaohui and Chen, Xinyuan and Wang, Limin and Lin, Dahua and Qiao, Yu and Liu, Ziwei},
  booktitle={IEEE/CVF Conference on Computer Vision and Pattern Recognition (CVPR)},
  pages={21807--21818},
  year={2024},
}

@article{lingbot2026,
  title={Scaling Mixture-of-Experts Video Pretraining for Embodied Intelligence},
  author={Ma, Shuailei and Liao, Jiaqi and Wang, Xinyang and Wang, Jingjing and Feng, Chaoran and Hu, Zijing and Bao, Chong and Xi, Zichen and Gan, Yuqi and Wang, Weisen and Zeng, Yanhong and Zhao, Qin and Shi, Zifan and Wu, Wei and Ouyang, Hao and Wang, Qiuyu and Zhang, Shangzhan and Shao, Jiahao and Sun, Yipengjing and Hu, Liangxiao and Pan, Lunke and Xue, Nan and Zheng, Kecheng and Xu, Yinghao and Zhu, Xing and Shen, Yujun and Cheng, Ka Leong},
  journal={arXiv preprint arXiv:2607.07675},
  year={2026}
}

@article{deepseekv3,
  title={{DeepSeek-V3} Technical Report},
  author={{DeepSeek-AI}},
  journal={arXiv preprint arXiv:2412.19437},
  year={2024}
}

@misc{magi2,
  title={{MAGI-2} Preview: Scaling Video Generation Models Efficiently},
  author={{Sand.ai}},
  year={2026},
  howpublished={Sand.ai Blog},
  url={https://sand.ai/blog/magi-2-preview}
}

@article{moepruner2024,
  title={{MoE-Pruner}: Pruning Mixture-of-Experts Large Language Model using the Hints from Its Router},
  author={Xie, Yanyue and Zhang, Zhi and Zhou, Ding and Xie, Cong and Song, Ziang and Liu, Xin and Wang, Yanzhi and Lin, Xue and Xu, An},
  journal={arXiv preprint arXiv:2410.12013},
  year={2024}
}

@inproceedings{reap2026,
  title={{REAP} the Experts: Why Pruning Prevails for One-Shot {MoE} Compression},
  author={Lasby, Mike and Lazarevich, Ivan and Sinnadurai, Nish and Lie, Sean and Ioannou, Yani and Thangarasa, Vithursan},
  booktitle={International Conference on Learning Representations (ICLR)},
  year={2026},
}

@inproceedings{mone2025,
  title={{MoNE}: Replacing Redundant Experts with Lightweight Novices for Structured Pruning of {MoE}},
  author={Zhang, Geng and Han, Yuxuan and Lou, Yuxuan and Zhang, Yiqi and Zhao, Wangbo and You, Yang},
  booktitle={International Conference on Learning Representations (ICLR)},
  year={2026},
}

@inproceedings{diep2025,
  title={{DiEP}: Adaptive Mixture-of-Experts Compression through Differentiable Expert Pruning},
  author={Bai, Sikai and Li, Haoxi and Zhang, Jie and Hong, Zicong and Guo, Song},
  booktitle={Advances in Neural Information Processing Systems},
  year={2025},
  doi={10.52202/085713-1878}
}

@article{evoesap2026,
  title={{EvoESAP}: Non-Uniform Expert Pruning for Sparse {MoE}},
  author={Liu, Zongfang and Tang, Shengkun and Sun, Boyang and Shen, Zhiqiang and Yuan, Xin},
  journal={arXiv preprint arXiv:2603.06003},
  year={2026}
}

@article{lexi2025,
  title={{LExI}: Layer-Adaptive Active Experts for Efficient {MoE} Model Inference},
  author={Chitty-Venkata, Krishna Teja and Madireddy, Sandeep and Emani, Murali and Vishwanath, Venkatram},
  journal={arXiv preprint arXiv:2509.02753},
  year={2025}
}

@article{ream2026,
  title={{REAM}: Merging Improves Pruning of Experts in {LLMs}},
  author={Jha, Saurav and Hashemzadeh, Maryam and Saheb Pasand, Ali and Parviz, Ali and Lee, Min-Joong and Knyazev, Boris},
  journal={arXiv preprint arXiv:2604.04356},
  year={2026}
}

@inproceedings{sere2026,
  title={{SERE}: Similarity-based Expert Re-routing for Efficient Batch Decoding in {MoE} Models},
  author={Wu, Juntong and Cheng, Jialiang and Lv, Fuyu and Dan, Ou and Yuan, Li},
  booktitle={International Conference on Learning Representations (ICLR)},
  year={2026},
  url={https://openreview.net/forum?id=98IxaUQtMY}
}

@inproceedings{shapleymoe2025,
  title={Discovering Important Experts for Mixture-of-Experts Models Pruning Through a Theoretical Perspective},
  author={Huang, Weizhong and Zhang, Yuxin and Zheng, Xiawu and Chao, Fei and Ji, Rongrong and Cao, Liujuan},
  booktitle={Advances in Neural Information Processing Systems},
  year={2025},
}

@article{shape2026,
  title={{SHAPE}: Coalition-Aware Expert Pruning for Sparse Mixture-of-Experts {LLMs}},
  author={Zhang, Yuhao},
  journal={arXiv preprint arXiv:2606.09886},
  year={2026}
}

@article{hts2026,
  title={How to Score Experts for One-Shot {MoE} Expert Pruning: A Unified Formulation and Selection Principle},
  author={Liu, Zongfang and Zhang, Jinghui and Ma, Zijian and Chen, Guangyi and Yuan, Xin},
  journal={arXiv preprint arXiv:2606.15716},
  year={2026}
}

@article{tbcoverage2026,
  title={Generic Expert Coverage for Pruning Sparse Mixture-of-Experts Language Models},
  author={Zeng, Yongqin and Pan, Sicheng and Wang, Jiale and Zheng, Hai-Tao and Kim, Hong-Gee and Ma, Chunxia and Zhou, Xiuteng},
  journal={arXiv preprint arXiv:2607.01710},
  year={2026}
}

@article{pathfinder2025,
  title={{MoE} Pathfinder: Trajectory-driven Expert Pruning},
  author={Yang, Xican and Tian, Yuanhe and Song, Yan},
  journal={arXiv preprint arXiv:2512.18425},
  year={2025}
}

@article{jacobs1991adaptive,
  title={Adaptive Mixtures of Local Experts},
  author={Jacobs, Robert A. and Jordan, Michael I. and Nowlan, Steven J. and Hinton, Geoffrey E.},
  journal={Neural Computation},
  volume={3}, number={1}, pages={79--87}, year={1991}
}

@article{jordan1994hierarchical,
  title={Hierarchical Mixtures of Experts and the {EM} Algorithm},
  author={Jordan, Michael I. and Jacobs, Robert A.},
  journal={Neural Computation}, volume={6}, number={2}, pages={181--214}, year={1994}
}

@inproceedings{lepikhin2021gshard,
  title={{GShard}: Scaling Giant Models with Conditional Computation and Automatic Sharding},
  author={Lepikhin, Dmitry and Lee, HyoukJoong and Xu, Yuanzhong and Chen, Dehao and Firat, Orhan and Huang, Yanping and Krikun, Maxim and Shazeer, Noam and Chen, Zhifeng},
  booktitle={International Conference on Learning Representations (ICLR)}, year={2021}
}

@inproceedings{du2022glam,
  title={{GLaM}: Efficient Scaling of Language Models with Mixture-of-Experts},
  author={Du, Nan and Huang, Yanping and Dai, Andrew M. and Tong, Simon and Lepikhin, Dmitry and Xu, Yuanzhong and Krikun, Maxim and Zhou, Yanqi and Yu, Adams Wei and Firat, Orhan and Zoph, Barret and Fedus, Liam and Bosma, Maarten P. and Zhou, Zongwei and Wang, Tao and Wang, Emma and Webster, Kellie and Pellat, Marie and Robinson, Kevin and Meier-Hellstern, Kathleen and Duke, Toju and Dixon, Lucas and Zhang, Kun and Le, Quoc and Wu, Yonghui and Chen, Zhifeng and Cui, Claire},
  booktitle={Proceedings of the 39th International Conference on Machine Learning},
  volume={162},
  pages={5547--5569},
  year={2022},
}

@article{jiang2024mixtral,
  title={Mixtral of Experts},
  author={Jiang, Albert Q. and Sablayrolles, Alexandre and Roux, Antoine and Mensch, Arthur and Savary, Blanche and Bamford, Chris and Chaplot, Devendra Singh and de las Casas, Diego and Bou Hanna, Emma and Bressand, Florian and Lengyel, Gianna and Bour, Guillaume and Lample, Guillaume and Renard Lavaud, L{\'e}lio and Saulnier, Lucile and Lachaux, Marie-Anne and Stock, Pierre and Subramanian, Sandeep and Yang, Sophia and Antoniak, Szymon and Le Scao, Teven and Gervet, Th{\'e}ophile and Lavril, Thibaut and Wang, Thomas and Lacroix, Timoth{\'e}e and El Sayed, William},
  journal={arXiv preprint arXiv:2401.04088}, year={2024}
}

@inproceedings{dai2024deepseekmoe,
  title={{DeepSeekMoE}: Towards Ultimate Expert Specialization in Mixture-of-Experts Language Models},
  author={Dai, Damai and Deng, Chengqi and Zhao, Chenggang and Xu, R.X. and Gao, Huazuo and Chen, Deli and Li, Jiashi and Zeng, Wangding and Yu, Xingkai and Wu, Y. and Xie, Zhenda and Li, Y.K. and Huang, Panpan and Luo, Fuli and Ruan, Chong and Sui, Zhifang and Liang, Wenfeng},
  booktitle={Proceedings of the 62nd Annual Meeting of the Association for Computational Linguistics (Volume 1: Long Papers)},
  pages={1280--1297},
  year={2024},
  publisher={Association for Computational Linguistics},
  doi={10.18653/v1/2024.acl-long.70}
}

@inproceedings{groeneveld2024olmoe,
  title={{OLMoE}: Open Mixture-of-Experts Language Models},
  author={Muennighoff, Niklas and Soldaini, Luca and Groeneveld, Dirk and Lo, Kyle and Morrison, Jacob and Min, Sewon and Shi, Weijia and Walsh, Pete and Tafjord, Oyvind and Lambert, Nathan and Gu, Yuling and Arora, Shane and Bhagia, Akshita and Schwenk, Dustin and Wadden, David and Wettig, Alexander and Hui, Binyuan and Dettmers, Tim and Kiela, Douwe and Farhadi, Ali and Smith, Noah A. and Koh, Pang Wei and Singh, Amanpreet and Hajishirzi, Hannaneh},
  booktitle={International Conference on Learning Representations (ICLR)},
  year={2025},
  url={https://openreview.net/forum?id=xXTkbTBmqq}
}

@inproceedings{clark2022unified,
  title={Unified Scaling Laws for Routed Language Models},
  author={Clark, Aidan and de las Casas, Diego and Guy, Aurelia and Mensch, Arthur and Paganini, Michela and Hoffmann, Jordan and Damoc, Bogdan and Hechtman, Blake and Cai, Trevor and Borgeaud, Sebastian and van den Driessche, George and Rutherford, Eliza and Hennigan, Tom and Johnson, Matthew and Millican, Katie and Cassirer, Albin and Jones, Chris and Buchatskaya, Elena and Budden, David and Sifre, Laurent and Osindero, Simon and Vinyals, Oriol and Rae, Jack and Elsen, Erich and Kavukcuoglu, Koray and Simonyan, Karen},
  booktitle={International Conference on Machine Learning (ICML)}, year={2022}
}

@inproceedings{peebles2023dit,
  title={Scalable Diffusion Models with Transformers},
  author={Peebles, William and Xie, Saining},
  booktitle={IEEE/CVF International Conference on Computer Vision (ICCV)}, year={2023}
}

@inproceedings{ho2022video,
  title={Video Diffusion Models},
  author={Ho, Jonathan and Salimans, Tim and Gritsenko, Alexey and Chan, William and Norouzi, Mohammad and Fleet, David J.},
  booktitle={Advances in Neural Information Processing Systems (NeurIPS)}, year={2022}
}

@article{blattmann2023svd,
  title={Stable Video Diffusion: Scaling Latent Video Diffusion Models to Large Datasets},
  author={Blattmann, Andreas and Dockhorn, Tim and Kulal, Sumith and Mendelevitch, Daniel and Kilian, Maciej and Lorenz, Dominik and Levi, Yam and English, Zion and Voleti, Vikram and Letts, Adam and Jampani, Varun and Rombach, Robin},
  journal={arXiv preprint arXiv:2311.15127}, year={2023}
}

@misc{brooks2024sora,
  title={Video Generation Models as World Simulators},
  author={{OpenAI}},
  year={2024},
  howpublished={OpenAI Technical Report},
  url={https://openai.com/index/video-generation-models-as-world-simulators/}
}

@inproceedings{han2015learning,
  title={Learning both Weights and Connections for Efficient Neural Networks},
  author={Han, Song and Pool, Jeff and Tran, John and Dally, William J.},
  booktitle={Advances in Neural Information Processing Systems (NeurIPS)}, year={2015}
}

@inproceedings{lecun1990optimal,
  title={Optimal Brain Damage},
  author={LeCun, Yann and Denker, John S. and Solla, Sara A.},
  booktitle={Advances in Neural Information Processing Systems},
  volume={2},
  pages={598--605},
  year={1989}
}

@inproceedings{hassibi1993second,
  title={Second Order Derivatives for Network Pruning: Optimal Brain Surgeon},
  author={Hassibi, Babak and Stork, David G.},
  booktitle={Advances in Neural Information Processing Systems},
  volume={5},
  pages={164--171},
  year={1992}
}

@inproceedings{li2016pruning,
  title={Pruning Filters for Efficient {ConvNets}},
  author={Li, Hao and Kadav, Asim and Durdanovic, Igor and Samet, Hanan and Graf, Hans Peter},
  booktitle={International Conference on Learning Representations (ICLR)}, year={2017}
}

@inproceedings{frankle2019lottery,
  title={The Lottery Ticket Hypothesis: Finding Sparse, Trainable Neural Networks},
  author={Frankle, Jonathan and Carbin, Michael},
  booktitle={International Conference on Learning Representations (ICLR)}, year={2019}
}

@inproceedings{ma2023llmpruner,
  title={{LLM-Pruner}: On the Structural Pruning of Large Language Models},
  author={Ma, Xinyin and Fang, Gongfan and Wang, Xinchao},
  booktitle={Advances in Neural Information Processing Systems (NeurIPS)}, year={2023}
}

@inproceedings{xia2024sheared,
  title={Sheared {LLaMA}: Accelerating Language Model Pre-training via Structured Pruning},
  author={Xia, Mengzhou and Gao, Tianyu and Zeng, Zhiyuan and Chen, Danqi},
  booktitle={International Conference on Learning Representations (ICLR)}, year={2024}
}

@article{ditmoe2024,
  title={Scaling Diffusion Transformers to 16 Billion Parameters},
  author={Zhengcong Fei and Mingyuan Fan and Changqian Yu and Debang Li and Junshi Huang},
  journal={arXiv preprint arXiv:2407.11633},
  year={2024}
}

@article{seedance2.0,
  title={Seedance 2.0: Advancing Video Generation for World Complexity},
  author={{Team Seedance} and Chen, De and Chen, Liyang and others},
  journal={arXiv preprint arXiv:2604.14148},
  year={2026}
}

@article{hope2026,
  title = {Higher-Order Pruning of Experts in Mixture-of-Experts Language Models},
  author = {Tseng, Alex M. and Kaul, Prannay and Zancato, Luca and Xia, Wei and Soatto, Stefano},
  journal = {arXiv preprint arXiv:2609.18916},
  year = {2026}
}

@article{coherentoverlap2026,
  title = {Beyond Geometric Complementarity: Coherent Overlap in Sparse Mixture-of-Experts Routing},
  author = {Tian, Huiyuan and Xu, Bonan and Li, Shijian},
  journal = {arXiv preprint arXiv:2607.28308},
  year = {2026}
}

@inproceedings{lu-etal-2024-experts,
    title = "Not All Experts are Equal: Efficient Expert Pruning and Skipping for Mixture-of-Experts Large Language Models",
    author = "Lu, Xudong  and
      Liu, Qi  and
      Xu, Yuhui  and
      Zhou, Aojun  and
      Huang, Siyuan  and
      Zhang, Bo  and
      Yan, Junchi  and
      Li, Hongsheng",
    editor = "Ku, Lun-Wei  and
      Martins, Andre  and
      Srikumar, Vivek",
    booktitle = "Proceedings of the 62nd Annual Meeting of the Association for Computational Linguistics (Volume 1: Long Papers)",
    month = aug,
    year = "2024",
    address = "Bangkok, Thailand",
    publisher = "Association for Computational Linguistics",
    doi = "10.18653/v1/2024.acl-long.334",
    pages = "6159--6172",
}
\bibliographystyle{arxiv_diet}

\newpage
\appendix

\setcounter{figure}{0}
\renewcommand{\thefigure}{A\arabic{figure}}
\setcounter{table}{0}
\renewcommand{\thetable}{A\arabic{table}}

\renewcommand{\topfraction}{0.92}
\renewcommand{\bottomfraction}{0.75}
\renewcommand{\textfraction}{0.08}
\renewcommand{\floatpagefraction}{0.72}
\setcounter{topnumber}{3}
\setcounter{bottomnumber}{2}
\setcounter{totalnumber}{5}

\makeatletter
\setlength{\@fptop}{0pt}
\setlength{\@fpbot}{0pt plus 1fil}
\makeatother

\section{Appendix Overview}
\label{app:overview}

This appendix provides additional analysis and experiments related to \ours{}, including:
\begin{itemize}
    \item Model architecture and grouped-routing formulation (Sec.~\ref{app:model});
    \item Calibration, video synthesis and benchmark protocols (Sec.~\ref{app:protocol});
    \item End-to-end compute budget (Sec.~\ref{app:compute});
    \item Regression-guided budget search details (Sec.~\ref{app:budgetdetails});
    \item The full layer-budget allocation sweep (Sec.~\ref{app:budget});
    \item Structural properties of the selected mask (Sec.~\ref{app:mask});
    \item Extended qualitative video comparisons (Sec.~\ref{app:qual});
    \item Routing concentration and activation entropy (Sec.~\ref{app:routing});
    \item Robustness to the calibration-set size (Sec.~\ref{app:calibsize});
    \item Response-space geometry and objective analysis (Sec.~\ref{app:corr});
    \item Response geometry across architecture depth (Sec.~\ref{app:atlas});
    \item Implementation and porting of the LLM MoE baselines (Sec.~\ref{app:baselines});
    \item Resolution transferability (Sec.~\ref{app:resolution});
    \item Model slicing and compact-checkpoint assembly (Sec.~\ref{app:injection});
    \item Physical memory and deployment footprint (Sec.~\ref{app:compression});
    \item Additional component ablations (Sec.~\ref{app:ablation});
    \item Paired statistical significance analysis (Sec.~\ref{app:paired});
    \item Geometric formulations and functional redundancy (Sec.~\ref{app:theory}).
\end{itemize}

\section{Architectural Formulation and Grouped Routing}
\label{app:model}

Consider a video diffusion transformer comprising $L$ blocks, where block $\ell$ incorporates a sparsely gated mixture-of-experts feed-forward layer with an expert bank $[E]=\{1,\dots,E\}$. For an incoming token representation $x\in\mathbb{R}^d$, the router evaluates routing logits and applies an element-wise sigmoid activation to yield unnormalized routing affinities $r_j(x)$. Under grouped routing, experts are partitioned into $G$ disjoint groups. Each group is scored by the sum of its two largest expert affinities; the top $G'$ scoring groups are activated, and the $k$ experts possessing the highest individual affinities within these selected groups are dispatched:
\begin{equation}
    y(x) \;=\; \sum_{j \in \mathcal{T}_k(x)} g_j(x)\, f_j(x),
    \qquad
    g_j(x) \;=\; \rho\,\frac{r_j(x)}{\sum_{i\in\mathcal{T}_k(x)} r_i(x)},
    \label{eq:routing}
\end{equation}
where $\mathcal{T}_k(x)\subseteq[E]$ denotes the index set of the $k$ active experts, $f_j(x)$ represents the output transformation of expert $j$, and $g_j(x)$ is the normalized gating weight scaled by the fixed route scale $\rho\approx1.25$ (the units of the $1.25$ gate weight in Appendix). In our backbone architecture, $L=48$, $E=128$, $G=4$, $G'=2$, and $k=8$ per layer.

Expert pruning identifies a pruned subset $\mathcal{D} \subset [E]$ and retains the complementary bank $\mathcal{K} = [E]\setminus\mathcal{D}$. For layer $\ell$, we denote the retained subset as $\mathcal{K}_\ell$ with retention count $k_\ell = |\mathcal{K}_\ell|$. Post-pruning inference executes the standard routing protocol, with pruned expert logits masked to $-\infty$ prior to grouped top-$k$ selection, followed by gate re-normalization over the retained experts. Crucially, pruning does not correspond to removing additive terms from a fixed linear sum; rather, it induces discrete token re-routing and reallocates gating weight among surviving experts. Scoring heuristics that treat experts independently of this dynamic redistribution fail to reflect runtime layer behavior.

\section{Experimental and Calibration Protocols}
\label{app:protocol}

\paragraph{Calibration capture.} 
We run a single instrumented calibration sweep over the unpruned pretrained model's $120$ calibration cases, recording routing logits, discrete routing assignments, gating weights, and complete expert output tensors across all layers and tokens. The calibration set encompasses $C=120$ video generation scenarios, tracking $64$ matched conditional and unconditional token pairs per scenario per layer ($368{,}640$ paired tokens, totaling $737{,}280$ branch tokens). The deletion-replay path is served by approximately $375$\,GB of cached expert state on disk in \texttt{fp16} precision (\texttt{bf16} gating values; this is the measured size of the compiled per-case caches, consistent with the raw tensor volume implied by the counts above), and the derived per-expert signature matrices add $2.9$\,GB; the two sizes differ by pooling, not precision. While evaluating all $128$ candidate experts per token incurs $\approx 16\times$ the active compute of a standard sparse forward pass, this overhead is front-loaded and incurred strictly once. All subsequent counterfactuals, including all $6{,}144$ single-expert deletions and arbitrary joint deletions, are executed via direct tensor operations on cached states (token states, router scores, grouped routing, gate weights and the expert output table, one file per case and layer) with zero additional model forward evaluations.

\paragraph{Video synthesis.} 
Videos are synthesized at a resolution of $240\times416$ across $121$ frames, $40$ denoising steps, classifier-free guidance scale $w=3$, flow-shift parameter $3$, and $24$\,fps. To ensure consistent evaluation, routing bias correction is disabled across all comparative experiments; validating on a matched $72$-case $480$p subset confirms that router bias state has negligible impact ($0.7916$ with bias enabled vs.\ $0.7923$ disabled). Every evaluation prompt is generated with a fixed random seed shared across all methods, ensuring strictly paired comparisons.

\paragraph{Benchmark evaluation.} 
We employ the official \vbench{} scorer across all $16$ standardized visual and semantic dimensions with official normalizations. To make full-scale comparative studies computationally feasible, we evaluate all models on a fixed $284$-case protocol. Calibration prompts are strictly excluded from the evaluation protocol, and missing pairs are never imputed.

\section{Compute Budget}
\label{app:compute}

The pipeline cost is front-loaded and dominated by two components. The all-expert calibration sweep over $120$ cases is the largest single expense: $\approx 12$ GPU-minutes per case on one GPU (per-case caches are then compiled on CPU), $\approx 24$ GPU-hours in total. The inter-layer search then consumed $87$ complete end-to-end evaluations of the $142$-case budget-search set, each $\approx 4$--$5$ GPU-hours, for $\approx 4\times10^{2}$ GPU-hours in total. Per-mask combinatorial optimization (GRASP initialization, single-swap local search and simulated annealing over the per-layer retention curves), together with the surrogate fit and the constrained linear-program allocation solve, runs entirely on CPU, within $\approx 1.5$--$2$ CPU-hours per mask. After calibration, every counterfactual, including all $6{,}144$ single-expert deletions and the joint-deletion probes used during selection, replays on cached states without additional model evaluations: \ours{} trades one all-expert capture and an offline search for the thousands of repeated forward passes that exhaustive leave-one-out simulation would require.

\section{Regression-Guided Layer Budget Search}
\label{app:budgetdetails}

The inter-layer budget allocation is optimized iteratively over a $142$-case budget-search set drawn from the prompt universe of the $284$-case evaluation suite. The search corpus comprises $87$ completed end-to-end evaluation trajectories across candidate retention allocations (excluding regimes beyond $80\%$ retention). Each iteration proceeds through three sequential stages:

\paragraph{Surrogate regression.} 
A regularized linear surrogate ($\lambda=1$, fitted with the exact closed form of Section~\ref{sec:budget}) maps the $48$-dimensional layer retention count vector $\mathbf{k} = [k_1, \dots, k_{48}]$ to measured dimension-wise benchmark scores. For surrogate fitting only, selection-corpus dimension scores are robustified by iterative Grubbs filtering; all reported benchmark results use the unfiltered aggregate over completed cases. With $87$ full trajectories supervising $48$ layer covariates, the surrogate operates in a data-constrained regime and serves as a coarse filter that proposes promising candidate allocations. Every proposed candidate is strictly validated via full end-to-end synthesis and official scoring, and the deployed allocation is the verified best among the evaluated trajectories. 

The evaluation suite splits into $142$ prompts shared with the budget-search set and $142$ strictly held-out prompts. Our selected allocation exhibits robust generalization: relative to the unpruned baseline, the paired difference is $+0.0200$ ($95\%$ CI $[+0.0018,+0.0429]$, $n=139$) on the shared partition and $+0.0142$ ($95\%$ CI $[-0.0008,+0.0298]$, $n=138$) on the held-out partition, fully aligning with the full-manifest improvement of $+0.0176$.

\begin{table}[!ht]
\centering
\caption{Comparative performance partitioned across the $284$-case protocol: the $142$ cases overlapping the search prompt distribution and the $142$ held-out cases unseen during search. Evaluation adheres to official \vbench{} normalization ($n=139$--$142$ completed pairs per cell).}
\label{tab:halves}
\small
\setlength{\tabcolsep}{7pt}
\begin{tabular}{lrr}
\toprule
Method & Shared $142$ & Held-Out $142$ \\
\midrule
Full Model (Unpruned) & 0.8023 & 0.7858 \\
Router Frequency & 0.7598 & 0.7367 \\
REAP~\citep{reap2026} & 0.7784 & 0.7745 \\
SHAPE~\citep{shape2026} & 0.7363 & 0.7569 \\
HTS~\citep{hts2026} & 0.7785 & 0.7858 \\
TB-Coverage~\citep{tbcoverage2026} & 0.7788 & 0.7637 \\
\textbf{\ours{} (Ours)} & \textbf{0.8222} & \textbf{0.8001} \\
\bottomrule
\end{tabular}
\end{table}

As documented in Table~\ref{tab:halves}, \ours{} maintains superior generative fidelity across both partitions, supporting transfer beyond the budget-search set. Every reported number is the plain aggregate over completed cases (an outlier-filtered variant was dropped).

\paragraph{Constrained linear-program proposal.} 
Total scores are linearized through the fixed normalization mapping. A constrained linear program proposes the allocation maximizing the predicted Total subject to the global budget constraint $\sum_\ell k_\ell = B$, architectural feasibility bounds, a one-sided per-dimension floor that bounds predicted regressions relative to the unpruned baseline (relaxable dimensions receive a soft penalty), and a Total-preservation constraint enforced as a hard floor at $50\%$ retention and above (softened to a deficit penalty below). The continuous optimum is integerized by a small mixed-integer step; the proposed allocation is then materialized, executed end-to-end, and appended to the search corpus.

The search follows a predefined schedule across retention ratios $\{0.5, 0.4, 0.6, 0.3, 0.7, 0.2, 0.8\}$. The optimal $50\%$ configuration retains between $52$ and $79$ experts per layer, establishing the allocation deployed for all comparative baselines in the main paper.

Leave-one-out surrogate predictions track empirical Total scores without systematic distortion, and the trajectory analysis confirms that the selected $50\%$ configuration defines the Pareto-optimal frontier.

\section{Layer-Budget Allocation Sweep}
\label{app:budget}

While the global retention budget is fixed, cross-layer allocation provides a critical degree of freedom.

\begin{wraptable}[13]{r}{0.53\textwidth}
\centering
\scriptsize
\renewcommand{\arraystretch}{0.95}
\begin{tabular}{crrr}
\toprule
Retention & Retained & Total & Kept Range \\
\midrule
$20\%$ & $1{,}229$ & 0.7710 & 16--116 \\
$30\%$ & $1{,}843$ & 0.8016 & 18--124 \\
$40\%$ & $2{,}458$ & 0.8015 & 16--128 \\
$50\%$ & $3{,}072$ & \textbf{0.8229} & 52--79 \\
$60\%$ & $3{,}686$ & 0.8103 & 64--94 \\
$70\%$ & $4{,}301$ & 0.8039 & 33--128 \\
$80\%$ & $4{,}915$ & 0.8195 & 26--126 \\
\bottomrule
\end{tabular}
\par\vspace{2pt}
{\footnotesize\captionof{table}{Exploration sweep on the $142$-case budget-search set; Total aggregates all completed cases of each run (selection-time scores).}\label{tab:sweep}}
\end{wraptable}

\begin{figure}[!ht]
\centering
\includegraphics[width=1.0\textwidth]{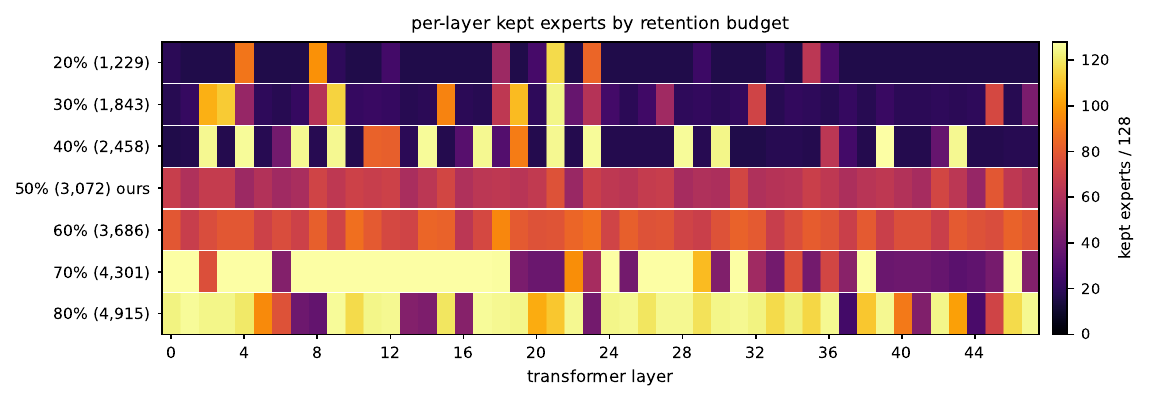}
\vspace{-0.10in}
\caption{Layer-wise retained expert counts across exploration budgets. Colors denote retaining counts out of $128$ candidate experts. The selected $50\%$ mask allocates capacity in a balanced band ($52$--$79$ experts per layer), whereas unconstrained exploration budgets assign as few as $16$ experts to resilient layers.}
\vspace{-0.15in}
\label{fig:layerbudget}
\end{figure}

Table~\ref{tab:sweep} reports the best-performing configurations identified across retention budgets on the budget-search set, and Figure~\ref{fig:layerbudget} visualizes the corresponding layer allocations.

\begin{figure}[!ht]
\centering
\includegraphics[width=0.98\textwidth]{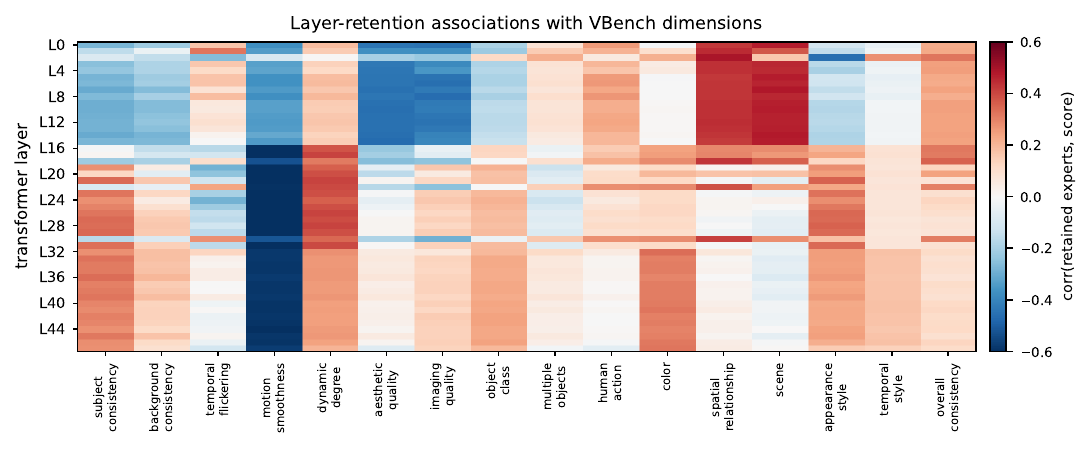}
\vspace{-0.10in}
\caption{Empirical correlation between layer-wise retained expert counts and standardized \vbench{} dimension scores across search trajectories, highlighting functional division of labor across network depth.}
\vspace{-0.15in}
\label{fig:layerdim}
\end{figure}

Optimized allocation interacts intimately with retained set selection criteria: the identified allocation vector confers a $1.2$-point end-to-end gain over uniform allocation for \ours{} (Section~\ref{sec:ablation}). Layer-wise budget allocation has been studied for language MoEs~\citep{diep2025,evoesap2026,lexi2025}; here allocation is searched jointly with the response-based selection from the same frozen capture and every proposed allocation is validated end to end, systematically discovering layer-specific capacity tolerances. The empirical search corpus enables both the found optimal across budgets (Table~\ref{tab:sweep}) and the cross-layer functional correlation map (Figure~\ref{fig:layerdim} and Section~\ref{sec:analysis}).

The sweep is not monotone: Total peaks at the $50\%$ budget, dips through $60$--$70\%$, and recovers only as retention approaches the unpruned model. The pruned $50\%$ allocation is also the most uniform (a $52$--$79$ per-layer range, against $64$--$94$ at $60\%$ and ranges spanning $16$--$128$ at the outer budgets), matching the search's preference for balanced allocations at the peak.

\section{Structural Properties of the Selected Mask}
\label{app:mask}

\begin{figure}[!ht]
\centering
\includegraphics[width=1.0\textwidth]{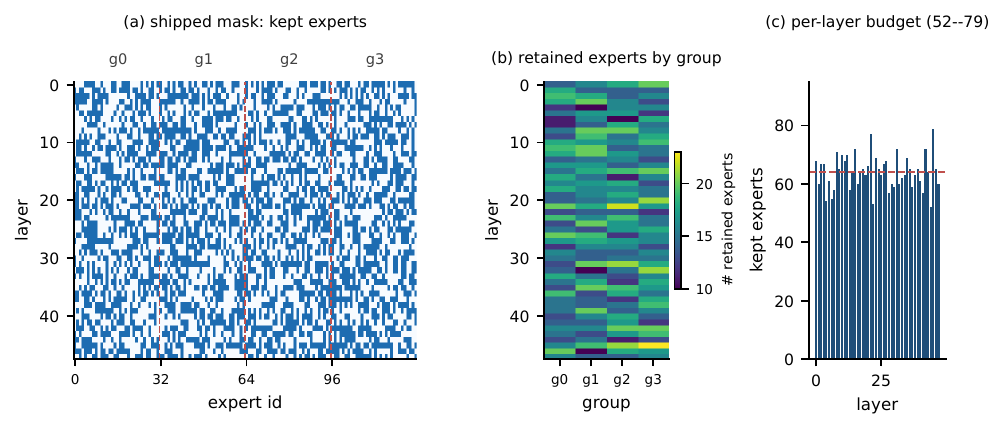}
\vspace{-0.10in}
\caption{Structural characteristics of the selected $50\%$ mask. (a) Full $48\times128$ retaining allocation pattern (dashed lines indicate group boundaries). (b) Retained counts per group per layer. (c) Total retained experts per layer. Every group retains at least $10$ experts and each layer preserves at least $52$, strictly satisfying grouped top-$8$ routing feasibility.}
\vspace{-0.15in}
\label{fig:mask}
\end{figure}

The selected $50\%$ mask satisfies architectural routing constraints with substantial margin (Figure~\ref{fig:mask}). Across all $48$ layers and four expert groups, the minimum retained expert count in any individual (layer, group) block is $10$ (well above the runtime threshold of four required for grouped top-$8$ routing), with a group-wise mean of $16.0$ retained experts. Excised experts are distributed evenly across all four routing groups rather than concentrating within specific subsets. Layer retention counts fluctuate smoothly between $52$ and $79$, capturing layer-specific sensitivity without inducing architectural bottlenecks.

\section{Extended Qualitative Comparisons}
\label{app:qual}

\begin{figure}[!ht]
\centering
\includegraphics[width=1.0\textwidth]{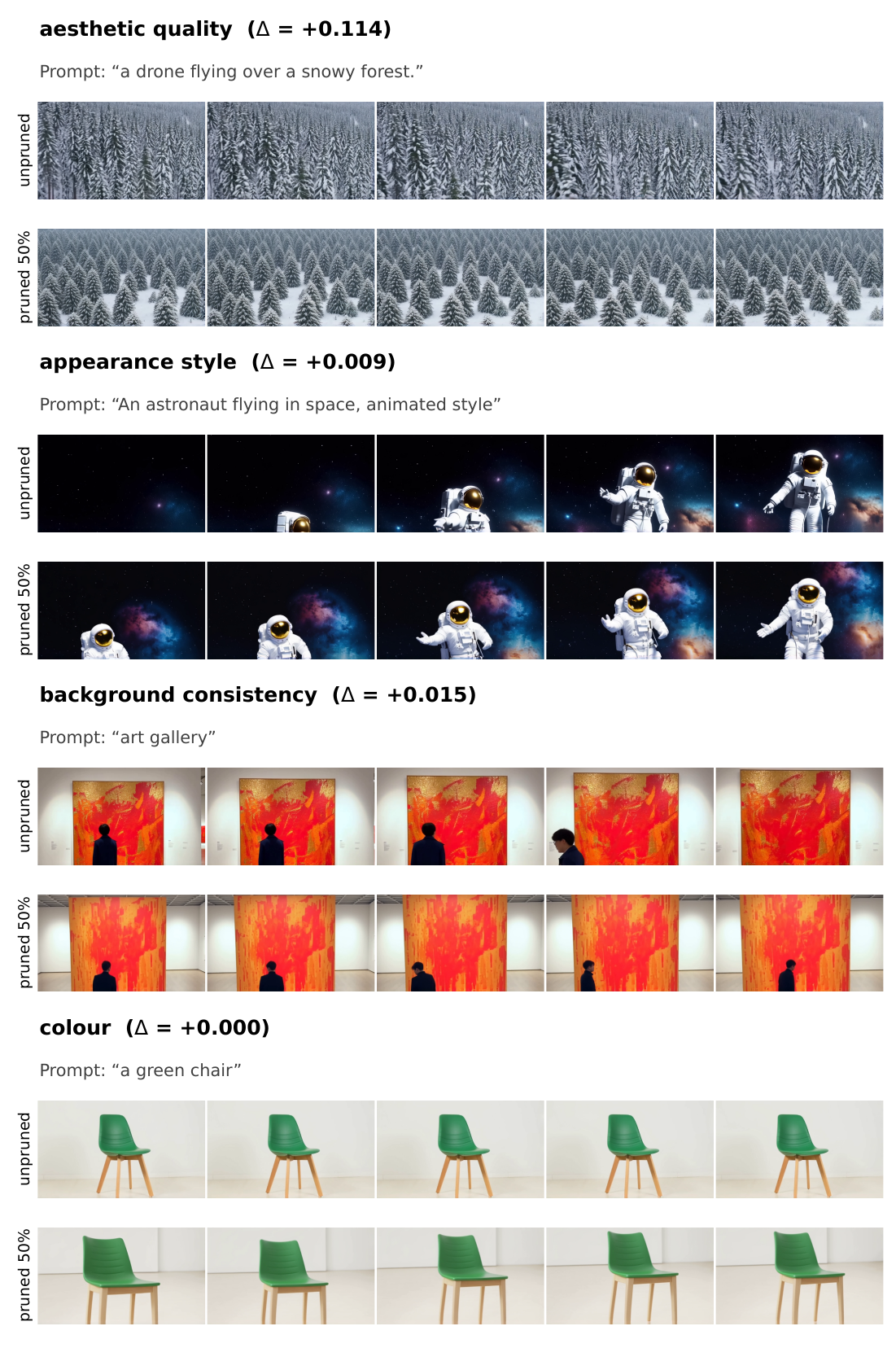}
\vspace{-0.10in}
\caption{Qualitative comparisons across visual appearance, background, and color fidelity dimensions. Each panel depicts representative generations under identical initializations for the dense baseline (top) and \ours{} at $50\%$ retention (bottom).}
\vspace{-0.15in}
\label{fig:qualitative_appendix}
\end{figure}

\begin{figure}[!ht]
\centering
\includegraphics[width=1.0\textwidth]{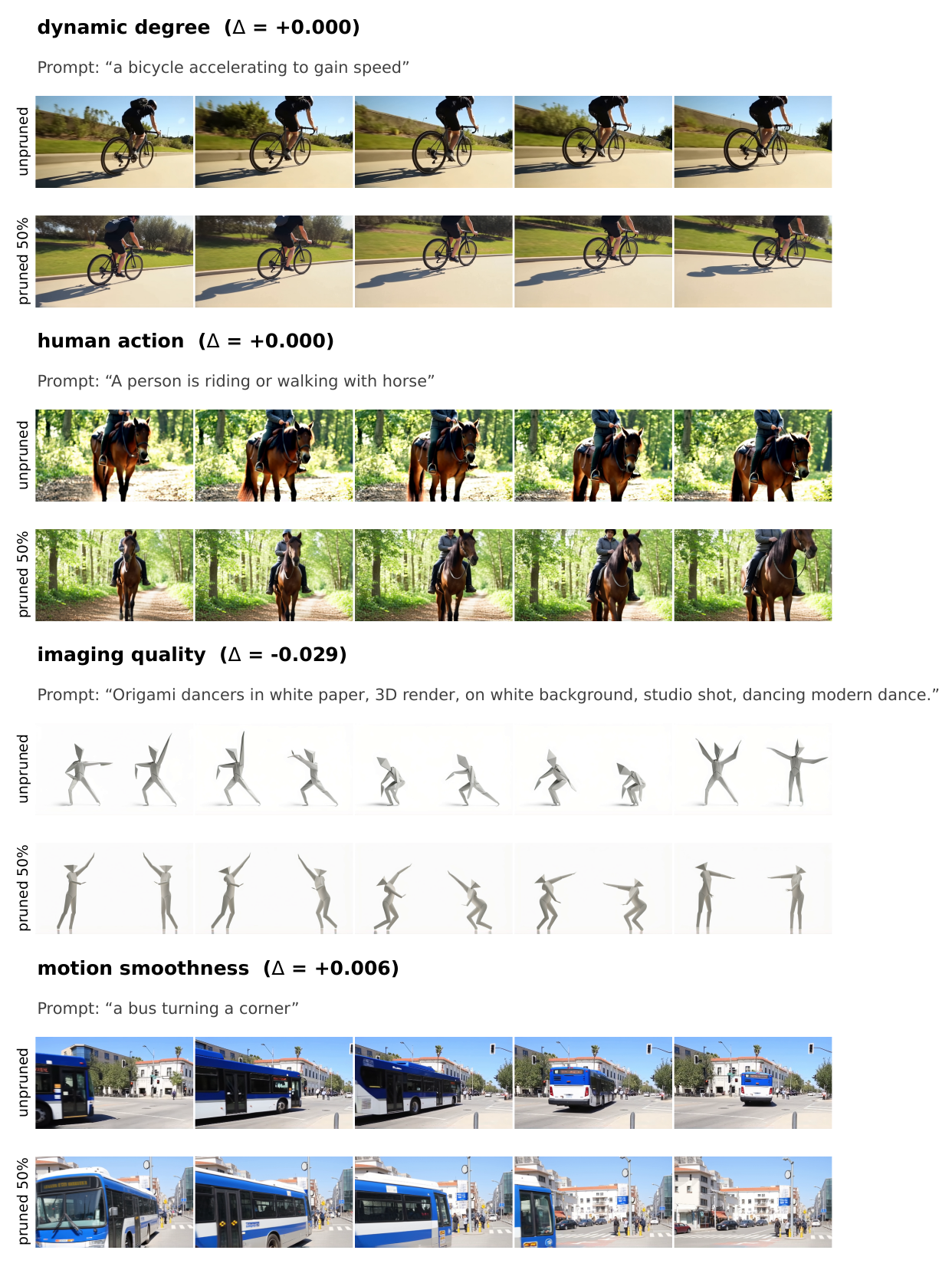}
\vspace{-0.10in}
\caption{Qualitative comparisons across dynamic degree, human action, imaging quality, and motion smoothness (protocol matches Figure~\ref{fig:qualitative_appendix}).}
\vspace{-0.15in}
\label{fig:gallery2}
\end{figure}

\begin{figure}[!ht]
\centering
\includegraphics[width=1.0\textwidth]{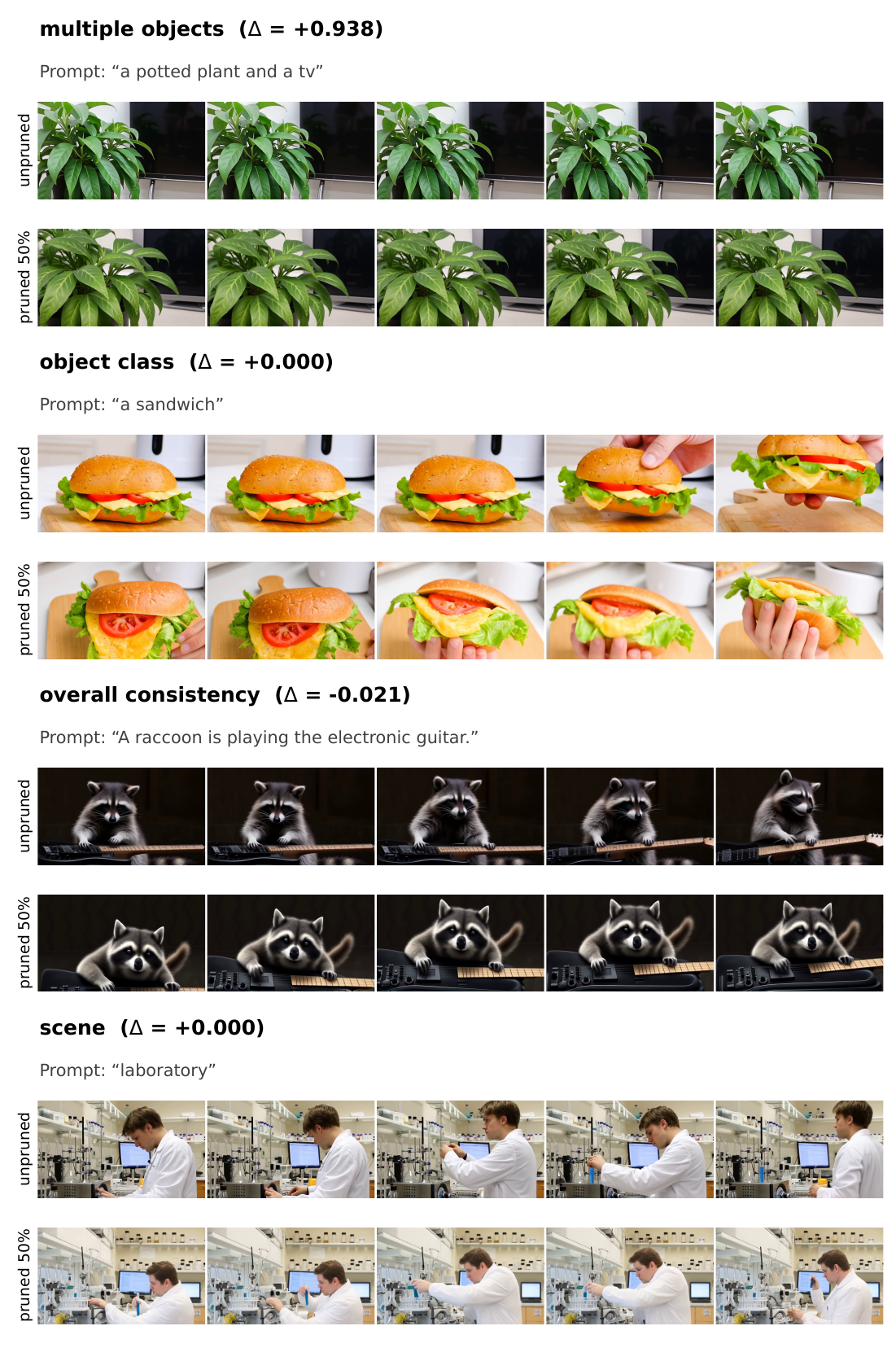}
\vspace{-0.10in}
\caption{Qualitative comparisons across multiple objects, object class, overall consistency, and scene composition (protocol matches Figure~\ref{fig:qualitative_appendix}).}
\vspace{-0.15in}
\label{fig:gallery3}
\end{figure}

\begin{figure}[!ht]
\centering
\includegraphics[width=1.0\textwidth]{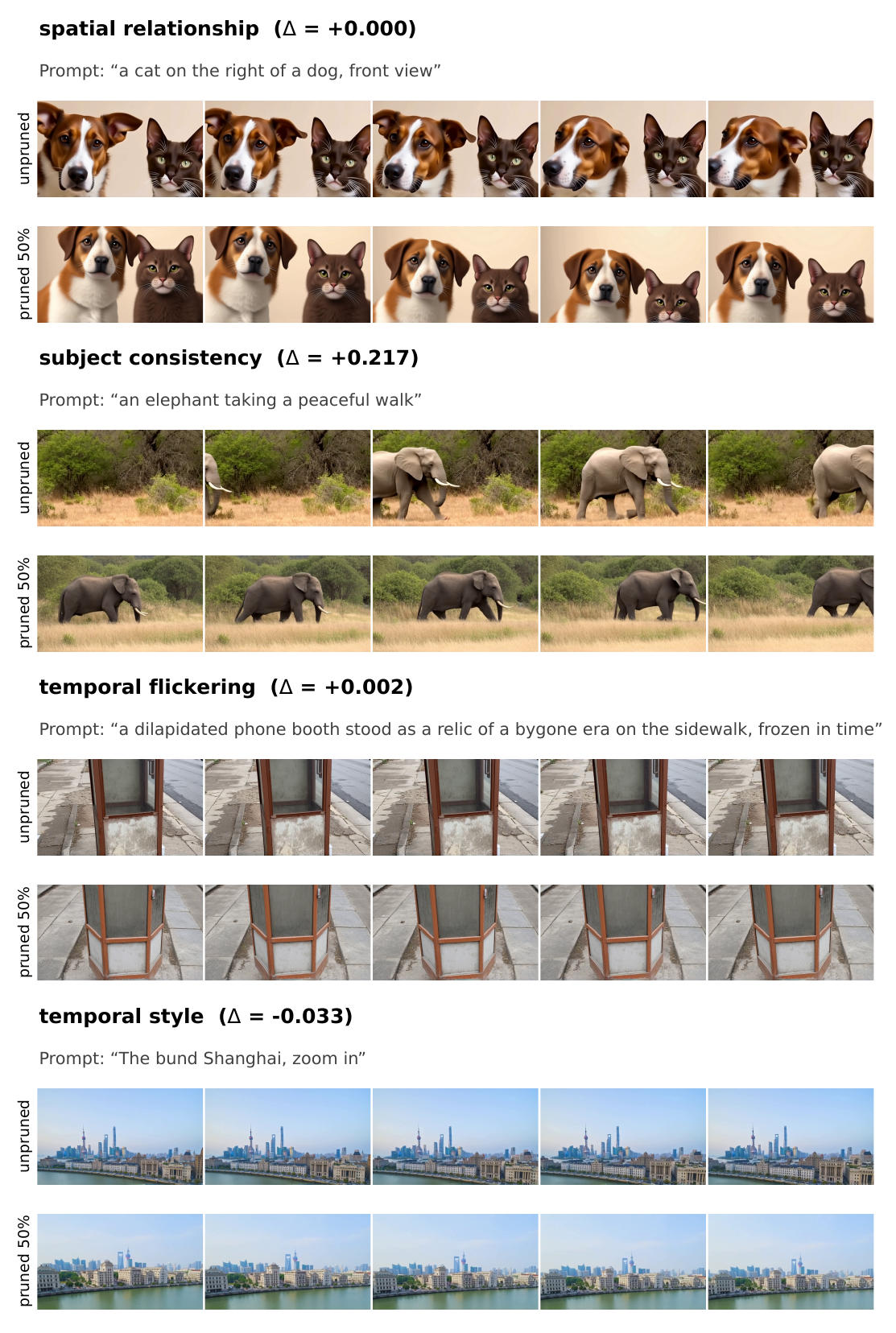}
\vspace{-0.10in}
\caption{Qualitative comparisons across spatial relationship, subject consistency, temporal flickering, and temporal style (protocol matches Figure~\ref{fig:qualitative_appendix}).}
\vspace{-0.15in}
\label{fig:gallery4}
\end{figure}

\begin{figure}[!ht]
\centering
\includegraphics[width=1.0\textwidth]{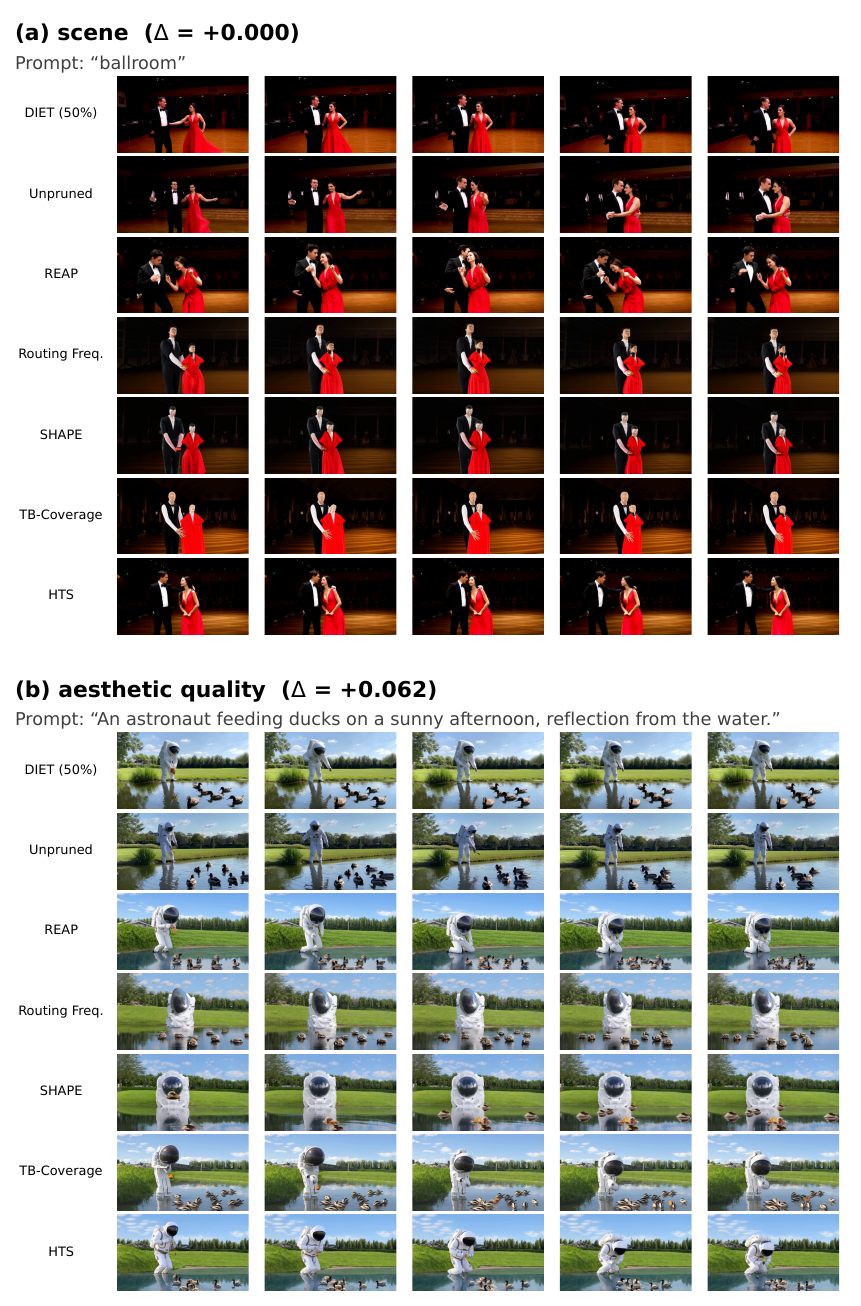}
\vspace{-0.10in}
\caption{All-arm frame-level comparisons on two representative cases: (a) a ballroom scene (``ballroom'') and (b) an aesthetic-quality prompt (``An astronaut feeding ducks on a sunny afternoon, reflection from the water.''). Each panel shows \ours{} at $50\%$ retention, the unpruned baseline, and the five ported criteria (named as in Table~\ref{tab:main}) as rows, with five evenly spaced frames of the same generation across columns. In (a) the scene composition is preserved across all arms, while in (b) the astronaut's suit collapses into a featureless mass under SHAPE and Routing Frequency; \ours{} remains visually consistent with the unpruned model in both panels.}
\vspace{-0.15in}
\label{fig:galleryadd}
\end{figure}

Figures~\ref{fig:qualitative_appendix}--\ref{fig:gallery4} present qualitative comparisons across all $16$ \vbench{} dimensions. For each dimension, we illustrate five evenly spaced frames generated under identical prompts and random seeds, comparing the unpruned baseline (top row) against \ours{} at $50\%$ retention (bottom row). To provide an objective representation rather than cherry-picked highlights, most displayed instances are the cases whose paired score differentials lie closest to their dimension's median performance delta, and the remaining instances show representative content for their dimension. Across seven of the sixteen dimensions, the median case exhibits negligible visual divergence, confirming that the pruned model reliably preserves baseline visual motion and fidelity. Figure~\ref{fig:galleryadd} complements these grids with frame-level, all-arm comparisons on two representative cases.

\FloatBarrier
\section{Routing Concentration and Activation Entropy}
\label{app:routing}

\begin{figure}[!ht]
\centering
\includegraphics[width=0.92\textwidth]{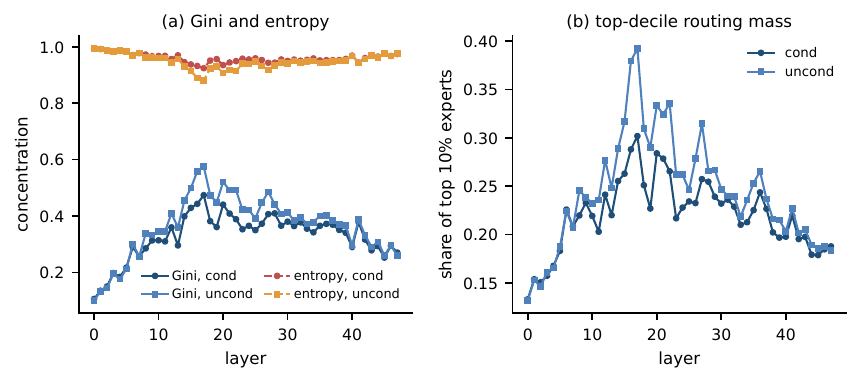}
\vspace{-0.10in}
\caption{Layer-wise routing concentration profiles across conditional and unconditional branches. (a) Gini coefficients and normalized entropy across depth. (b) Token share absorbed by the top-decile experts. Routing remains well-dispersed across all layers, demonstrating the absence of inactive capacity.}
\vspace{-0.15in}
\label{fig:routing}
\end{figure}

Both CFG branches exhibit smooth, depth-dependent dispersion (Figure~\ref{fig:routing}): the conditional branch displays an average Gini of $0.328$ and a top-decile token share of $0.220$, while the unconditional branch exhibits $0.364$ and $0.243$, with average normalized entropies of $0.960$ and $0.950$, respectively. Crucially, no layer exhibits collapsed or degenerate routing. Pruning based on usage frequency can excise moderately active units possessing unique functional roles, necessitating principled geometric coverage to prevent capacity loss.

\section{Robustness to Calibration Set Size}
\label{app:calibsize}

To verify that retained expert selection does not overfit to the calibration sample, we analyze solver stability as a function of calibration size $C$. We re-solve the combinatorial selection using identical algorithmic hyperparameters---four starts (one deterministic backward-greedy path and three randomized GRASP initializations), each refined through $50$ swap passes, $300\text{k}$ simulated annealing steps, and $50$ final swap iterations---evaluating signatures computed over subsets of $C \in \{15, 30, 60\}$ cases across five independent draws. 

\begin{figure}[!ht]
\centering
\includegraphics[width=0.98\textwidth]{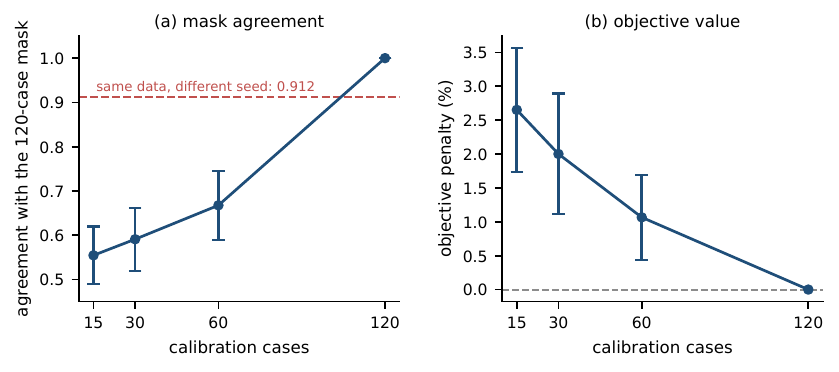}
\vspace{-0.10in}
\caption{Stability across calibration sample sizes. (a) Retained expert mask overlap between $C$-case solutions and the $120$-case reference (dashed line indicates multi-seed solver variance of $0.912$). (b) Relative ODL objective value evaluated against the full calibration set. Reducing calibration size preserves functional objective quality within $2.7\%$.}
\vspace{-0.15in}
\label{fig:calibration}
\end{figure}

As shown in Figure~\ref{fig:calibration}, solution overlap with the $120$-case reference mask scales smoothly: $0.555\pm0.065$ ($C=15$), $0.591\pm0.071$ ($C=30$), and $0.668\pm0.078$ ($C=60$). When evaluated under the full $120$-case objective, masks derived from these smaller subsets achieve objective values within $2.7\%$, $2.0\%$, and $1.1\%$ of the best solution found on the 120-case objective, respectively.

This variation must be evaluated relative to stochastic solver variance. Executing the solver on the full $120$-case dataset across different random seeds yields an average mask overlap of $0.912$ (minimum $0.800$) with a negligible objective variation of $0.0003$. Because the optimization landscape is relatively flat near the optimum, reducing calibration size primarily shifts selection among functionally equivalent retained expert configurations rather than degrading functional coverage.

\FloatBarrier
\section{Analysis of Response-Space Geometry and Objectives}
\label{app:corr}

\begin{figure}[t]
\centering
\includegraphics[width=0.76\textwidth]{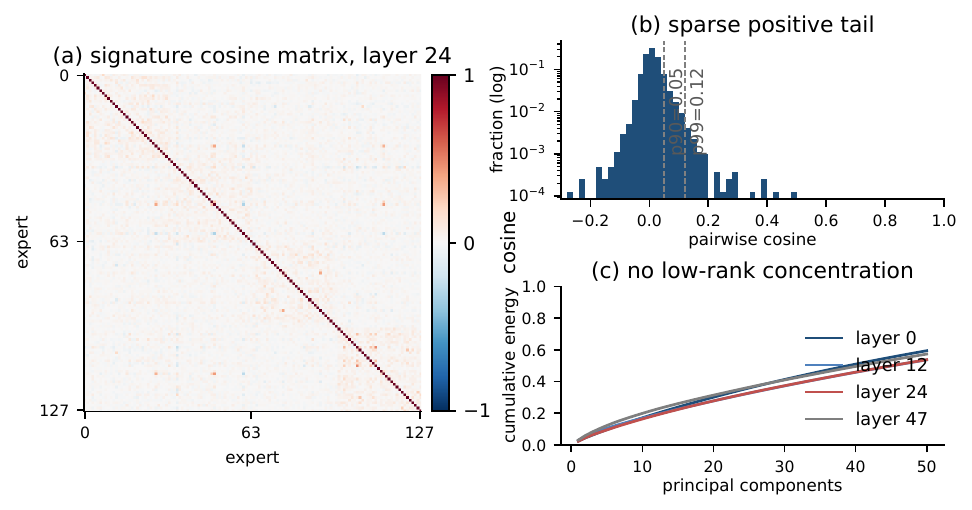}
\vspace{-0.12in}
\caption{Geometric properties of deletion-response signatures at layer $24$. (a) Pairwise signature cosine matrix. (b) Distribution of pairwise cosines (logarithmic scale), demonstrating sharp concentration near zero with an extended positive tail. (c) Cumulative singular value energy, exhibiting absence of low-rank concentration.}
\vspace{-0.16in}
\label{fig:geometry}
\end{figure}

Signatures exhibit near-orthogonality (mean pairwise cosine $0.008$--$0.019$) and exceptionally high dimensionality, with participation ratios spanning $93$--$119$ out of $128$ and a flat singular spectrum across all depths (Figures~\ref{fig:atlas}, \ref{fig:geometry}). This absence of low-rank subspace concentration renders classical projection or principal-component pruning ineffective; instead, the primary exploitable structure is a sparse tail of functionally aligned pairs, whose mask-level statistics are detailed below and in Table~\ref{tab:twins}.

The intrinsic geometry of deletion-response signatures dictates the performance bounds of candidate pruning rules. Across all layers, signatures exhibit near-orthogonality (mean pairwise cosine $0.008$--$0.019$) and high intrinsic dimensionality, with singular spectrum participation ratios spanning $93$--$119$ out of $128$ and the top $50$ principal components accounting for only $54$--$60\%$ of cumulative variance (Figures~\ref{fig:geometry}, \ref{fig:atlas}). Consequently, subspace-based compression heuristics (e.g., volume maximization or principal component truncation) lack low-rank structure to exploit (Table~\ref{tab:subspace}).

Under this setting, the most salient exploitable structure is the sparse positive tail of the pairwise cosine distribution: while the median pairwise cosine is $0.01$, a minor subset of expert pairs exhibits marked functional redundancy ($\cos \approx 0.6$--$0.9$, with $<0.2\%$ of pairs exceeding $0.3$; Figure~\ref{fig:geometry}b). In a $50\%$ pruning regime ($\approx 64$ deletions per layer in an effective $\approx 100$-dimensional manifold), no retained expert subset can uniformly span all deleted directions. Pruning algorithms must therefore prioritize excising redundant directions. As confirmed in Table~\ref{tab:twins}, \ours{} systematically identifies these redundant pairs: pruned experts exhibit an average nearest retained expert cosine of $0.17$ (vs.\ $0.09$ among retained experts), with $94\%$ of deletions possessing a surviving substitute above $0.1$ cosine similarity.

We validate alternative objective formulations end-to-end under identical layer allocations ($52$--$79$ experts retained per layer, Table~\ref{tab:subspace}). \emph{D-optimal volume selection} maximizes $\log\det(\mathbf{C}_\mathcal{K})$, where $\mathbf{C}$ is the pairwise cosine kernel. \emph{PCA energy selection} retains experts exhibiting the highest projection energy along the top-$r$ principal eigenvectors ($r \in \{8, 32\}$). Variants operating in alternative feature spaces evaluate Equation~\ref{eq:odl} using uncontextualized expert outputs or routing logits.

Consistent with the flat eigenvalue spectrum, PCA energy pruning lags behind \ours{} by $2.66$--$2.78$ points ($0.7849$ and $0.7837$), degrading below the unpruned baseline. Rebuilding ODL over isolated expert outputs or routing affinities similarly incurs steep penalties ($0.7883$ and $0.7813$, trailing \ours{} by $+0.0234$ and $+0.0299$ paired delta, respectively), confirming that contextual interaction captured via deletion signatures is critical. D-optimal volume maximization attains $0.8007$ ($+0.0102$ paired difference, CI $[-0.0037,+0.0259]$); however, it isolates significantly fewer redundant tail pairs ($62$ vs.\ $135$ high-cosine deletions).

\begin{figure}[!ht]
\centering
\includegraphics[width=0.80\textwidth]{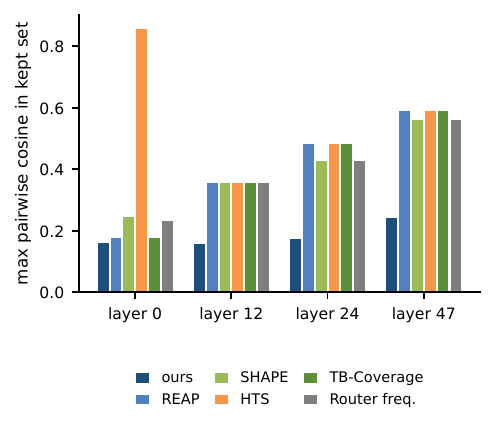}
\vspace{-0.10in}
\caption{Maximum pairwise signature cosine within the retained expert sets across four representative layers. \ours{} consistently maintains the lowest internal redundancy, preserving diverse directional coverage.}
\vspace{-0.15in}
\label{fig:redundancy}
\end{figure}

As detailed in Table~\ref{tab:twins}, while all methods incur an average per-expert cosine loss close to $1.0$ due to high dimensionality, \ours{} excises $135$ experts possessing close surviving substitutes ($\cos > 0.3$), compared to at most $26$ for baseline criteria. Furthermore, \ours{} retains the least internally redundant retained expert sets, bounding maximum pairwise retained expert cosine to $0.16$--$0.24$ across representative layers (Figure~\ref{fig:redundancy}).

\begin{table}[!ht]
\centering
\small
\setlength{\tabcolsep}{5pt}
\begin{tabular}{lccc}
\toprule
Selection Objective (Matched Budget) & Total ODL $\downarrow$ & High-$\cos$ Deletions $\uparrow$ & \vbench{} Total $\uparrow$ \\
\midrule
\textbf{\ours{} (ODL, Response Space)} & \textbf{2561.2} & \textbf{135} & \textbf{0.8115} \\
D-Optimal (Volume Maximization) & 2648.1 & 62 & 0.8007 \\
PCA Energy ($r=32$) & 2732.9 & 5 & 0.7849 \\
PCA Energy ($r=8$) & 2755.8 & 3 & 0.7837 \\
ODL (Isolated Output Space) & 2676.1 & 74 & 0.7883 \\
ODL (Router Score Space) & 2692.6 & 77 & 0.7813 \\
\bottomrule
\end{tabular}
\vspace{-0.06in}
\caption{End-to-end comparison of candidate selection objectives under identical layer budgets ($3{,}072$ retained experts). High-$\cos$ deletions denote pruned experts whose nearest retained expert exhibits $\cos > 0.3$. Low-rank PCA truncation and space-ablated variants degrade performance by $2.32$--$3.02$ points.}
\vspace{-0.10in}
\label{tab:subspace}
\end{table}

\begin{table}[!ht]
\centering
\small
\setlength{\tabcolsep}{4.5pt}
\begin{tabular}{lcccccc}
\toprule
Selection Rule & Total ODL & Mean Loss & Deletions & Deletions & Mean $\|\sig_e\|$ \\
& $\sum_e \delta_e \downarrow$ & Per Expert $\downarrow$ & ($\cos > 0.3$) $\uparrow$ & ($\cos > 0.5$) $\uparrow$ & of Excised \\
\midrule
\textbf{\ours{} ($50\%$)} & \textbf{2561.2} & \textbf{0.834} & \textbf{135} & \textbf{27} & 0.706 \\
HTS~\citep{hts2026} & 2760.4 & 0.899 & 23 & 2 & 0.521 \\
REAP~\citep{reap2026} & 2763.7 & 0.900 & 22 & 2 & 0.484 \\
SHAPE~\citep{shape2026} & 2767.4 & 0.901 & 26 & 1 & 0.468 \\
Router Frequency & 2769.7 & 0.902 & 21 & 1 & 0.463 \\
TB-Coverage~\citep{tbcoverage2026} & 2772.6 & 0.903 & 25 & 2 & 0.440 \\
\bottomrule
\end{tabular}
\vspace{-0.06in}
\caption{Deletion profiles across pruning heuristics under matched layer budgets ($3{,}072$ retained experts, $3{,}072$ deletions). Loss denotes $\delta_e = 1-\cos(\sig_e,\sig_{f(e)})$. \ours{} specifically targets functionally covered experts ($135$ with $\cos > 0.3$) rather than indiscriminately pruning low-norm units.}
\vspace{-0.10in}
\label{tab:twins}
\end{table}

\FloatBarrier

\section{Response Geometry Across Architecture Depth}
\label{app:atlas}

\begin{figure}[!ht]
\centering
\includegraphics[width=0.96\textwidth]{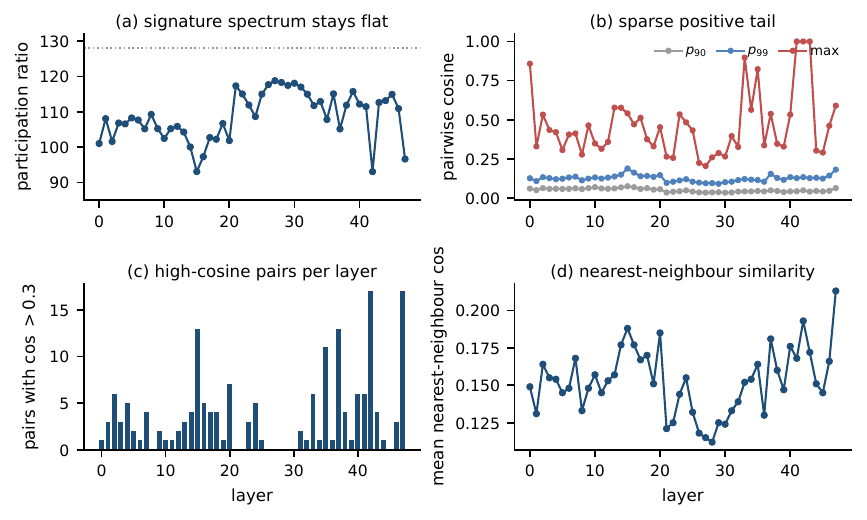}
\vspace{-0.10in}
\caption{Layer-wise geometric profile across all $48$ transformer blocks. (a) Singular spectrum participation ratios. (b) Pairwise cosine quantiles and maxima. (c) Density of high-cosine pairs ($\cos > 0.3$). (d) Mean nearest-neighbor cosine.}
\vspace{-0.15in}
\label{fig:atlas}
\end{figure}

The geometric characteristics observed at individual layers generalize throughout the network (Figure~\ref{fig:atlas}). The singular spectrum remains uniformly flat across all $48$ layers, with participation ratios averaging $109$ (range $93$--$119$). Pairwise cosines are densely concentrated around zero ($p_{90} \in [0.035, 0.076]$, $p_{99} \in [0.09, 0.19]$). While sparse positive tails exist across all depths, their density varies smoothly (averaging $3.6$ pairs with $\cos > 0.3$ per layer). Thus, near-orthogonality and sparse directional alignment represent intrinsic structural features of the video MoE backbone.

\FloatBarrier
\section{Implementation and Porting of LLM MoE Baselines}
\label{app:baselines}

\begin{table}[!ht]
\centering
\caption{Summary of evaluated MoE pruning baselines adapted to video DiT architectures~\citep{reap2026,shape2026,hts2026,tbcoverage2026,lu-etal-2024-experts}. All methods utilize identical calibration tensors, evaluation protocols, and budget allocations.}
\label{tab:baselines}
\small
\begin{tabular}{llll}
\toprule
Method & Pruning Metric & Source Domain & Reference \\
\midrule
Router Frequency & Token dispatch counts & LLM MoE & Standard Baseline \\
REAP & Gating weight $\times$ activation norm & LLM MoE & \citet{reap2026} \\
SHAPE & Coalition Shapley values & LLM MoE & \citet{shape2026} \\
HTS & Activation norm formulations & LLM MoE & \citet{hts2026} \\
TB-Coverage & Round-robin register protection & LLM MoE & \citet{tbcoverage2026} \\
NAEE-Style Recon & Replayed layer-output reconstruction & LLM MoE & \citet{lu-etal-2024-experts} \\
\midrule
\textbf{\ours{} (Ours)} & Deletion-response overall diversity loss & Video DiT & This work \\
\bottomrule
\end{tabular}
\end{table}

\paragraph{Porting baseline criteria.}
All ported baselines consume identical calibration tensors recorded across the $120$ calibration runs over all $48$ layers, executing under the same per-layer retention allocations as \ours{} (summarized in Table~\ref{tab:baselines}):
\begin{itemize}[leftmargin=1.2em,itemsep=1pt,topsep=2pt]
\item \textbf{Router Frequency.} Scores each expert by total routed token volume across matched conditional and unconditional branches.
\item \textbf{REAP.} Implements the routing-weighted activation saliency metric of~\citet{reap2026}:
\[
    s(e) \;=\; \frac{1}{|\mathcal{X}_e|}\sum_{t\,:\, e \in \mathcal{T}_k(x_t)} g_e(x_t)\,\|f_e(x_t)\|_2,
\]
where $|\mathcal{X}_e|$ is the total number of captured tokens dispatched to expert $e$.
\item \textbf{SHAPE.} Evaluates marginal Shapley values across observed expert coalitions~\citep{shape2026}, weighting marginal sub-coalition contributions by empirical occurrence frequency.
\item \textbf{HTS.} Ports the unified scoring family of~\citet{hts2026} and, for each retention budget, reports the stronger of its two task-agnostic gate-free criteria: Mean Squared Activation Norm (MSAN, $S(1,0,2)$), averaging $\|f_e(x_t)\|_2^2$, and Mean Activation Norm (MAN, $S(1,0,1)$), averaging $\|f_e(x_t)\|_2$, across routed tokens.
\item \textbf{TB-Coverage.} Ports the coverage pipeline of Generic Expert Coverage~\citep{tbcoverage2026}: each register scores experts with the REAP-style conditional-mean utility and ranks them independently, a round-robin protected set is built by alternating between the two rankings, the protected set is merged with reconstruction-stable candidates selected to minimize the fixed-gate output reconstruction error, and the lowest-scoring unprotected experts are removed to restore the budget. The video setting has no WikiText2/C4 text corpora, so the two registers are formed from the two prompt families of the calibration corpus (appearance/composition against temporal/physics), with the protection budget fixed at $B=16$, the largest value feasible at every layer.
\item \textbf{NAEE-Style Recon.} Minimizes the replayed layer-output difference~\citep{lu-etal-2024-experts}: for every layer, the solver of Section~\ref{sec:objective} is re-run with the objective replaced by the guidance-weighted $\ell_2$ difference between the replayed and the captured layer output, at the same budgets, and the selection receives the same grouped-feasibility repair as the other ports. Evaluated on the full $120$-case calibration set, this port reduces the mean replayed output error by about $40\%$ relative to \ours{} ($1.94$ to $1.96$ against $3.21$ to $3.22$ across the uniform and searched allocations).
\end{itemize}

All baseline selections are mapped to satisfy grouped-routing feasibility constraints; where raw selections violate the four-retained expert group floor, the highest-scoring feasible subset is retained. Complete dimension-wise results are compiled in Table~\ref{tab:perdim} ($50\%$); Tables~\ref{tab:perdim20} and~\ref{tab:perdim80} give the $20\%$ and $80\%$ budgets.

\begin{table}[!ht]
\centering
\caption{Dimension-wise \vbench{} evaluation on the full $284$-case protocol under matched $50\%$ budgets ($3{,}072$ retained experts).}
\label{tab:perdim}
\scriptsize
\setlength{\tabcolsep}{3pt}
\begin{tabular}{lrrrrrrrr}
\toprule
Dimension & Full model & Router freq. & REAP & SHAPE & HTS & TB-Coverage & NAEE-Style Recon & Ours \\
\midrule
Aesthetic qual. & 0.554 & 0.503 & 0.555 & 0.485 & 0.580 & 0.528 & 0.540 & 0.598 \\
Appearance style & 0.806 & 0.789 & 0.785 & 0.773 & 0.802 & 0.812 & 0.817 & 0.813 \\
Background cons. & 0.937 & 0.929 & 0.953 & 0.920 & 0.955 & 0.953 & 0.947 & 0.954 \\
Color & 0.845 & 0.868 & 0.675 & 0.768 & 0.738 & 0.794 & 0.905 & 0.809 \\
Dynamic degree & 0.733 & 0.467 & 0.600 & 0.600 & 0.733 & 0.600 & 0.733 & 0.800 \\
Human action & 1.000 & 0.905 & 0.952 & 0.905 & 0.905 & 0.952 & 0.952 & 1.000 \\
Imaging qual. & 0.572 & 0.582 & 0.565 & 0.548 & 0.584 & 0.573 & 0.565 & 0.579 \\
Motion smooth. & 0.973 & 0.984 & 0.969 & 0.983 & 0.969 & 0.981 & 0.974 & 0.981 \\
Multiple objects & 0.610 & 0.426 & 0.467 & 0.390 & 0.610 & 0.423 & 0.544 & 0.702 \\
Object class & 0.846 & 0.699 & 0.801 & 0.702 & 0.831 & 0.820 & 0.798 & 0.871 \\
Overall cons. & 0.740 & 0.658 & 0.694 & 0.661 & 0.691 & 0.679 & 0.627 & 0.723 \\
Scene & 0.359 & 0.190 & 0.384 & 0.279 & 0.296 & 0.283 & 0.194 & 0.389 \\
Spatial rel. & 0.628 & 0.501 & 0.446 & 0.525 & 0.460 & 0.463 & 0.519 & 0.639 \\
Subject cons. & 0.905 & 0.866 & 0.926 & 0.848 & 0.919 & 0.903 & 0.919 & 0.927 \\
Temporal flicker. & 0.974 & 0.975 & 0.983 & 0.980 & 0.978 & 0.979 & 0.983 & 0.972 \\
Temporal style & 0.650 & 0.551 & 0.628 & 0.566 & 0.667 & 0.577 & 0.573 & 0.605 \\
\bottomrule
\end{tabular}

\end{table}

\begin{table}[!ht]
\centering
\caption{Dimension-wise \vbench{} evaluation on the full $284$-case protocol at $20\%$ retention ($1{,}229$ retained experts).}
\label{tab:perdim20}
\scriptsize
\setlength{\tabcolsep}{3pt}
\begin{tabular}{lrrrrrrrr}
\toprule
Dimension & Full model & Router freq. & REAP & SHAPE & HTS & TB-Coverage & NAEE-Style Recon & Ours \\
\midrule
Aesthetic qual. & 0.554 & 0.336 & 0.423 & 0.355 & 0.415 & 0.423 & 0.436 & 0.535 \\
Appearance style & 0.806 & 0.762 & 0.762 & 0.767 & 0.787 & 0.757 & 0.771 & 0.804 \\
Background cons. & 0.937 & 0.932 & 0.967 & 0.931 & 0.957 & 0.958 & 0.925 & 0.935 \\
Color & 0.845 & 0.818 & 0.662 & 0.700 & 0.615 & 0.762 & 0.764 & 0.769 \\
Dynamic degree & 0.733 & 0.667 & 0.267 & 0.533 & 0.533 & 0.600 & 0.467 & 0.400 \\
Human action & 1.000 & 0.286 & 0.524 & 0.381 & 0.762 & 0.476 & 0.571 & 0.857 \\
Imaging qual. & 0.572 & 0.459 & 0.516 & 0.481 & 0.562 & 0.558 & 0.574 & 0.604 \\
Motion smooth. & 0.973 & 0.981 & 0.984 & 0.976 & 0.979 & 0.983 & 0.979 & 0.988 \\
Multiple objects & 0.610 & 0.026 & 0.136 & 0.055 & 0.254 & 0.103 & 0.195 & 0.507 \\
Object class & 0.846 & 0.346 & 0.449 & 0.298 & 0.412 & 0.511 & 0.471 & 0.801 \\
Overall cons. & 0.740 & 0.487 & 0.519 & 0.494 & 0.486 & 0.547 & 0.556 & 0.643 \\
Scene & 0.359 & 0.013 & 0.160 & 0.068 & 0.114 & 0.203 & 0.258 & 0.224 \\
Spatial rel. & 0.628 & 0.053 & 0.148 & 0.108 & 0.362 & 0.161 & 0.193 & 0.409 \\
Subject cons. & 0.905 & 0.875 & 0.953 & 0.872 & 0.941 & 0.944 & 0.892 & 0.927 \\
Temporal flicker. & 0.974 & 0.980 & 0.981 & 0.974 & 0.978 & 0.971 & 0.982 & 0.984 \\
Temporal style & 0.650 & 0.350 & 0.383 & 0.359 & 0.442 & 0.446 & 0.417 & 0.515 \\
\bottomrule
\end{tabular}

\end{table}

\begin{table}[!ht]
\centering
\caption{Dimension-wise \vbench{} evaluation on the full $284$-case protocol at $80\%$ retention ($4{,}915$ retained experts).}
\label{tab:perdim80}
\scriptsize
\setlength{\tabcolsep}{3pt}
\begin{tabular}{lrrrrrrrr}
\toprule
Dimension & Full model & Router freq. & REAP & SHAPE & HTS & TB-Coverage & NAEE-Style Recon & Ours \\
\midrule
Aesthetic qual. & 0.554 & 0.585 & 0.576 & 0.577 & 0.572 & 0.589 & 0.579 & 0.586 \\
Appearance style & 0.806 & 0.814 & 0.799 & 0.806 & 0.795 & 0.811 & 0.814 & 0.800 \\
Background cons. & 0.937 & 0.934 & 0.958 & 0.944 & 0.953 & 0.957 & 0.954 & 0.959 \\
Color & 0.845 & 0.797 & 0.769 & 0.808 & 0.783 & 0.794 & 0.861 & 0.790 \\
Dynamic degree & 0.733 & 0.800 & 0.867 & 0.800 & 0.867 & 0.867 & 0.867 & 0.867 \\
Human action & 1.000 & 1.000 & 1.000 & 1.000 & 1.000 & 1.000 & 0.952 & 1.000 \\
Imaging qual. & 0.572 & 0.571 & 0.567 & 0.571 & 0.572 & 0.553 & 0.567 & 0.570 \\
Motion smooth. & 0.973 & 0.967 & 0.957 & 0.967 & 0.959 & 0.956 & 0.957 & 0.967 \\
Multiple objects & 0.610 & 0.596 & 0.618 & 0.621 & 0.669 & 0.614 & 0.629 & 0.603 \\
Object class & 0.846 & 0.864 & 0.816 & 0.846 & 0.846 & 0.853 & 0.835 & 0.875 \\
Overall cons. & 0.740 & 0.739 & 0.741 & 0.738 & 0.740 & 0.728 & 0.734 & 0.738 \\
Scene & 0.359 & 0.253 & 0.313 & 0.355 & 0.279 & 0.380 & 0.287 & 0.304 \\
Spatial rel. & 0.628 & 0.601 & 0.524 & 0.496 & 0.459 & 0.553 & 0.562 & 0.513 \\
Subject cons. & 0.905 & 0.883 & 0.918 & 0.881 & 0.925 & 0.916 & 0.915 & 0.924 \\
Temporal flicker. & 0.974 & 0.978 & 0.979 & 0.968 & 0.973 & 0.973 & 0.969 & 0.975 \\
Temporal style & 0.650 & 0.674 & 0.668 & 0.672 & 0.673 & 0.656 & 0.663 & 0.659 \\
\bottomrule
\end{tabular}

\end{table}


\section{Resolution Transferability}
\label{app:resolution}

We evaluate the generalization of the pruning configuration to the backbone's native $480$p resolution ($480\times720$) across identical prompts and random seeds.

\begin{figure}[!ht]
\centering
\includegraphics[width=0.95\textwidth]{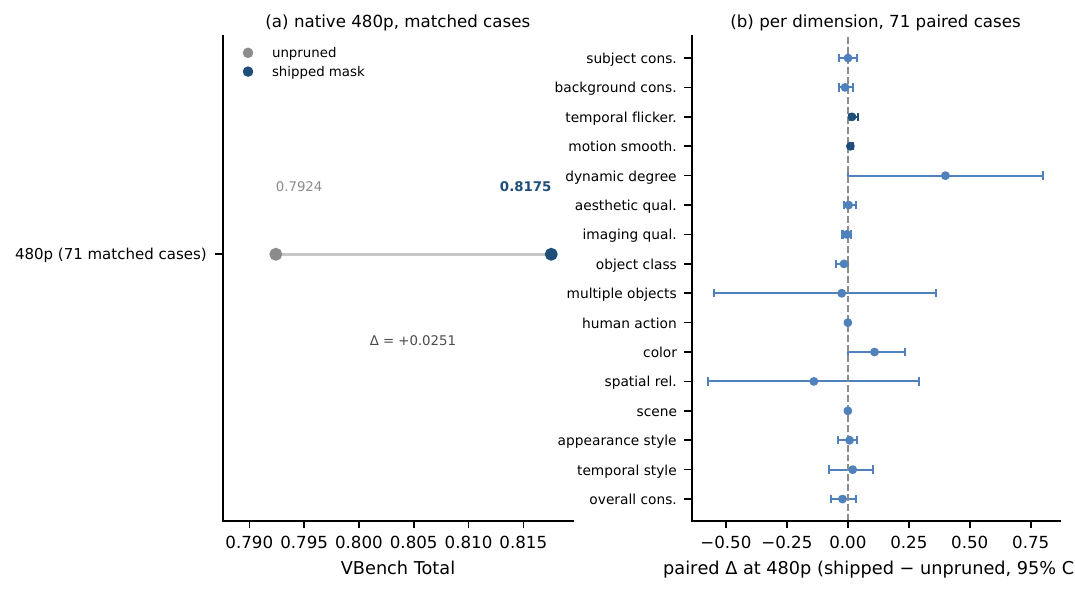}
\vspace{-0.10in}
\caption{Cross-resolution evaluation. (a) Total at native $480$p for the unpruned model and the pruned mask, on the $71$ matched cases of the resolution screen. (b) Paired dimension-wise differentials at native $480$p (pruned mask minus unpruned, $95\%$ CI).}
\vspace{-0.15in}
\label{fig:480p}
\end{figure}

On the $72$ cases shared by this screen and the $284$-case protocol: the unpruned model scores $0.7923$ at $480$p, and the pruned mask reaches $0.8175$ against $0.7924$ for the baseline on the $71$ cases that exclude the one prompt shared with calibration ($+0.0251$ paired, $95\%$ CI $[-0.0051, +0.0563]$, $n=71$; per-dimension shifts in Figure~\ref{fig:480p}b).

\begin{figure}[!ht]
\centering
\includegraphics[width=0.90\textwidth]{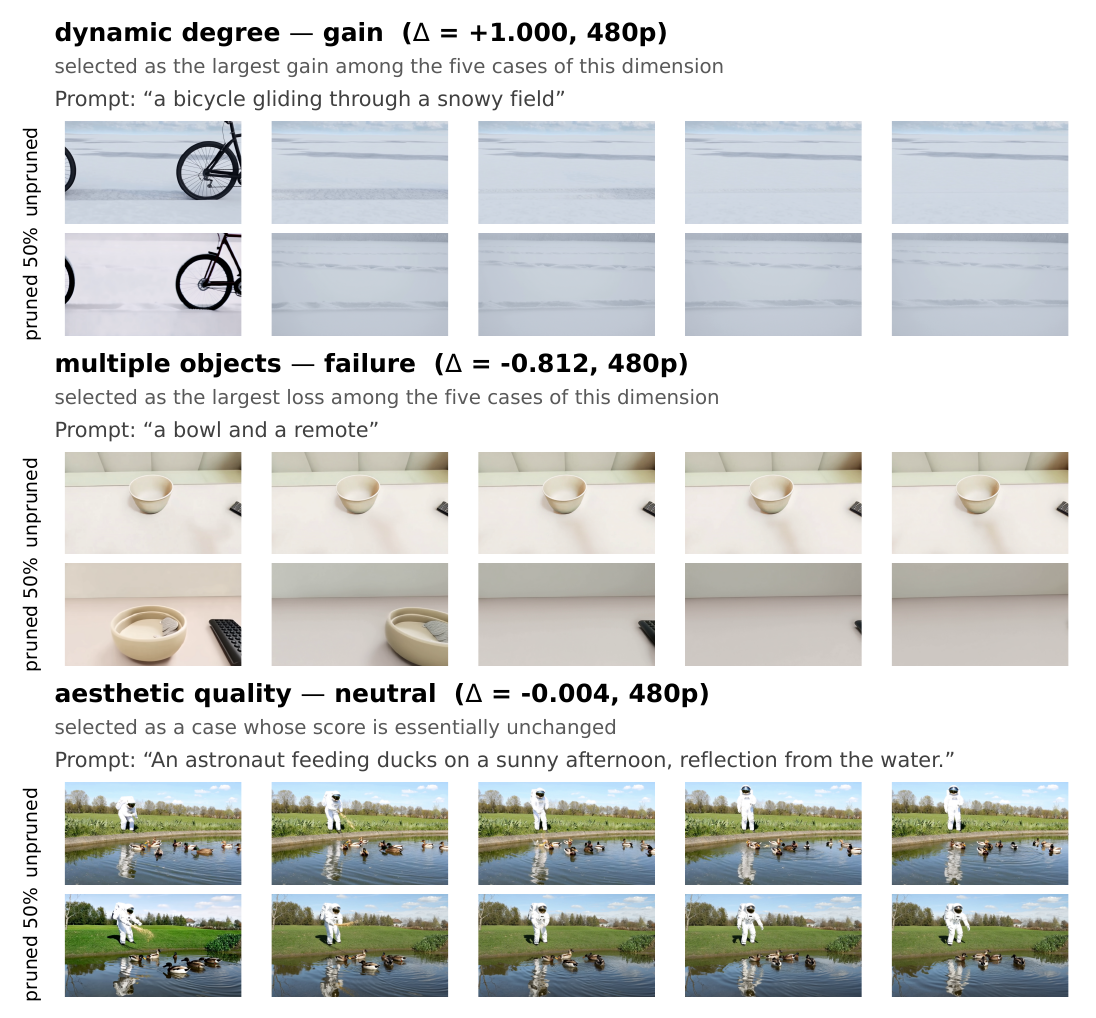}
\vspace{-0.10in}
\caption{Representative native $480$p generations across positive, neutral, and sensitive dimensions under matched initializations. Panels compare unpruned baselines (top row) against \ours{} at $50\%$ retention (bottom row).}
\vspace{-0.15in}
\label{fig:qual480p}
\end{figure}

To put the pruned checkpoint on the same footing, the mask of Section~\ref{sec:main} was run across the full $284$-case protocol at native $480$p on single-card replicas, attaining a Total score of $0.8179$ (compared to $0.8115$ for the same mask at $240$p). The two $480$p checks differ in mask, case set and run configuration, so their levels are not comparable; both agree that the pruned model holds at the higher resolution. The paired differential between resolutions is $+0.0066$ ($95\%$ CI $[-0.0087, +0.0203]$, $n=279$), with no systematic qualitative artifacts (Figure~\ref{fig:qual480p}).

\FloatBarrier
\section{Model Slicing and Compact Checkpoint Assembly}
\label{app:injection}

Retained expert masks can be deployed via either dynamic runtime masking or physical tensor compaction, both preserving exact routing semantics. In dynamic runtime masking, the router evaluates all $E$ candidate logits, masks excised expert indices to $-\infty$, and re-normalizes surviving gates. In physical compaction (utilized for all reported results), the expert weight tensors are physically sliced along the expert dimension, reducing the expert bank to $k_\ell$ surviving blocks. The routing module retains its original index projection, masking excised logits and utilizing an index lookup table to direct surviving indices to their compacted physical memory slots. Attention parameters, shared experts, and normalization layers are preserved unmodified.

Numerical verification confirms that the offline simulator and the compacted runtime model exhibit high-fidelity numerical agreement, displaying a mean token-wise relative output error of $3.4\times10^{-3}$ across all $48$ layers ($40$ evaluated scenarios). This residual error arises strictly from floating-point caching quantization (\texttt{bfloat16} gating values and \texttt{fp16} expert outputs): evaluating routing in \texttt{float64} shifts the output delta by $<10^{-9}$, whereas modifying routing logic to ungrouped top-$k$ increases output divergence to $0.23$. Thus, the optimization objective operates on a high-fidelity representation of the physical inference operator.

\FloatBarrier
\section{Physical Memory and Deployment Footprint}
\label{app:compression}

As detailed in Table~\ref{tab:compression}, while active computational cost per step remains constant, halving the expert bank reduces the replica's stored weights from $57$\,GB to $30$\,GB in \texttt{bfloat16}. Consequently, a complete model replica fits on a single $48$\,GB GPU (e.g., NVIDIA RTX 6000 Ada, A40, or L40), eliminating inter-GPU communication and halving deployment costs.

\begin{table}[!ht]
\centering
\caption{Physical deployment and memory footprint comparison between the unpruned model and \ours{} at $50\%$ retention. Measurements reflect empirical single-replica execution. Active FLOPs per token remain strictly identical.}
\label{tab:compression}
\small
\begin{tabular}{lrr}
\toprule
Metric & Full Model & \textbf{\ours{} ($50\%$)} \\
\midrule
Routed Expert Count & $6{,}144$ & $3{,}072$ \\
Expert Bank Parameters & $29.0\text{B}$ & $14.5\text{B}$ \\
Checkpoint Disk Footprint (\texttt{bfloat16}) & $57$\,GB & $30$\,GB \\
Required GPUs per Replica ($48$\,GB VRAM) & $2$ & \textbf{1} \\
Total Replica VRAM Allocation & $93.2$\,GiB ($2\times46.6$) & \textbf{44.7\,GiB} \\
Peak Memory ($240\times416$, Batch Size 2) & $46.6$\,GiB & $44.7$\,GiB \\
Active Experts per Token & $8$ & $8$ \\
Active FLOPs per Token & \multicolumn{2}{c}{Identical} \\
\bottomrule
\end{tabular}
\end{table}

\section{Additional Ablations}
\label{app:ablation}

\begin{figure}[!ht]
\centering
\includegraphics[width=0.62\textwidth]{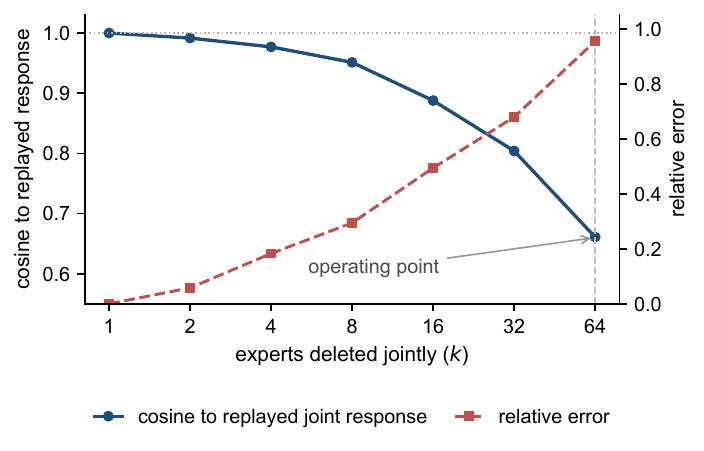}
\vspace{-0.10in}
\caption{Breakdown of additive superposition under joint multi-expert deletions. Cosine similarity between true joint response perturbations and constituent single-deletion sums degrades sharply with deletion cardinality.}
\vspace{-0.15in}
\label{fig:mechanism}
\end{figure}

Table~\ref{tab:ablation} summarizes ablation experiments isolating core framework components. Formulating selection via the summed-response norm---an additive proxy prevalent in classical pruning literature---precipitates a $5.44$-point degradation, collapsing spatial coherence dimensions from $0.63$ to $0.33$. As demonstrated in Figure~\ref{fig:mechanism}, this degradation is consistent with the observed non-additivity of multi-expert removals: the cosine similarity between a joint deletion perturbation and the linear sum of single-deletion responses degrades from $1.00$ at $k=1$ to $0.66$ at $k=64$, with relative reconstruction error escalating to $0.95$. Consequently, additive norm metrics fail to model multi-expert removal dynamics.

\section{Paired Statistical Significance Analysis}
\label{app:paired}

\begin{figure}[!ht]
\centering
\includegraphics[width=0.80\textwidth]{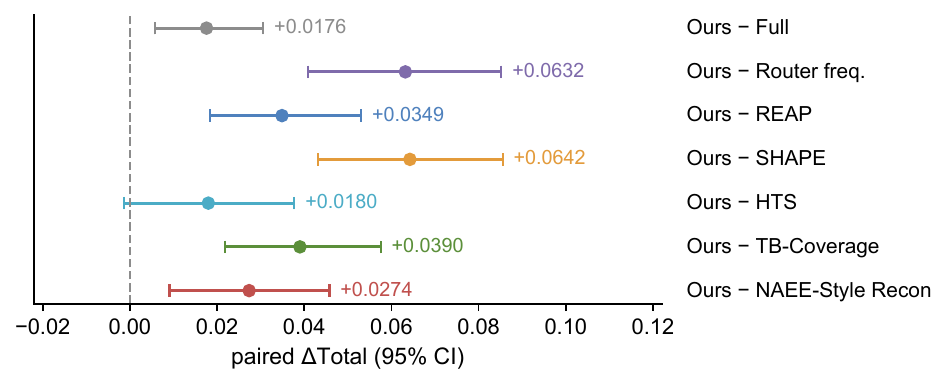}
\vspace{-0.10in}
\caption{Paired per-case bootstrap differentials ($95\%$ percentile intervals) comparing \ours{} against the dense baseline and ported MoE pruning heuristics at $50\%$ retention.}
\vspace{-0.15in}
\label{fig:pairedsummary}
\end{figure}

Because all evaluations share identical prompt sequences and seed initializations, statistical significance is quantified via paired bootstrap resampling ($10{,}000$ iterations) over prompt cases (Figure~\ref{fig:pairedsummary}).

Relative to the unpruned baseline, \ours{} achieves an overall paired differential of $\Delta\text{Total} = +0.0176$ ($95\%$ CI $[+0.0057, +0.0306]$, $n=277$). When dynamic degree is excluded and remaining dimensions are re-normalized, the improvement remains statistically significant ($+0.0145$, $95\%$ CI $[+0.0039, +0.0249]$, $n=262$), confirming that the aggregate gain is not solely driven by dynamic degree. Against ported LLM baselines, the paired differentials over common completed cases are statistically significant for Router Frequency $+0.0632$ ($95\%$ CI $[+0.0409, +0.0852]$), SHAPE $+0.0642$ ($95\%$ CI $[+0.0431, +0.0855]$), REAP $+0.0349$ ($95\%$ CI $[+0.0183, +0.0531]$), TB-Coverage $+0.0390$ ($95\%$ CI $[+0.0219, +0.0575]$), and NAEE-Style Recon $+0.0274$ ($95\%$ CI $[+0.0091, +0.0458]$); against HTS, the strongest ported baseline (best of its variants; Table~\ref{tab:main}), the paired differential is a positive but not statistically significant $+0.0180$ ($95\%$ CI $[-0.0013, +0.0377]$); note that this paired estimate over common completed cases differs from the direct mean difference of $0.8115-0.7921 = 0.0194$ over the full protocol.

\begin{figure}[!ht]
\centering
\includegraphics[width=0.94\textwidth]{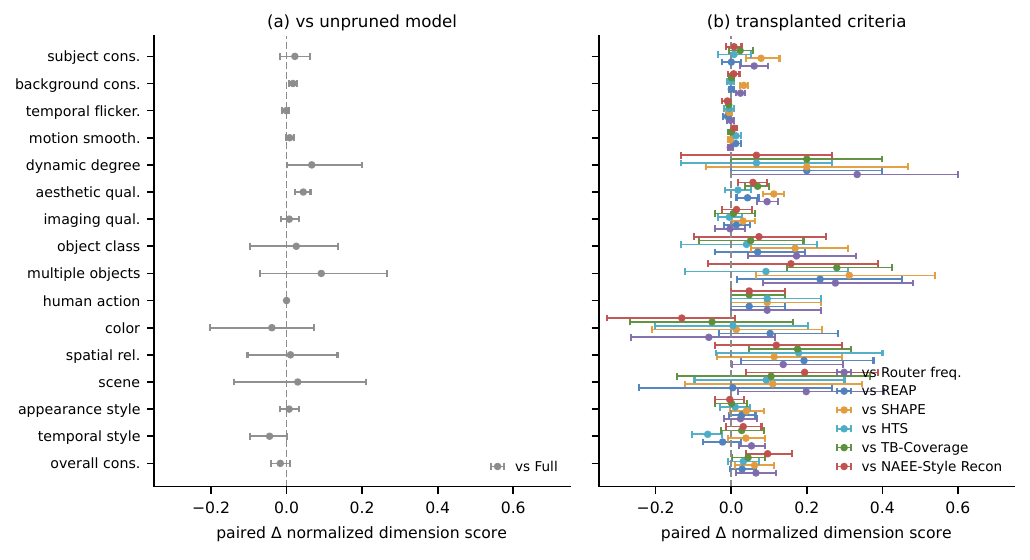}
\vspace{-0.10in}
\caption{Dimension-wise paired differentials ($95\%$ bootstrap intervals) comparing \ours{} against (a) the unpruned baseline and (b) ported LLM pruning baselines, including the reconstruction-criterion port NAEE-Style Recon. While differentials against the unpruned baseline exhibit conservative preservation across dimensions, comparisons against ported baselines demonstrate decisive semantic advantages.}
\vspace{-0.15in}
\label{fig:perdim}
\end{figure}

\begin{figure}[!ht]
\centering
\includegraphics[width=0.94\textwidth]{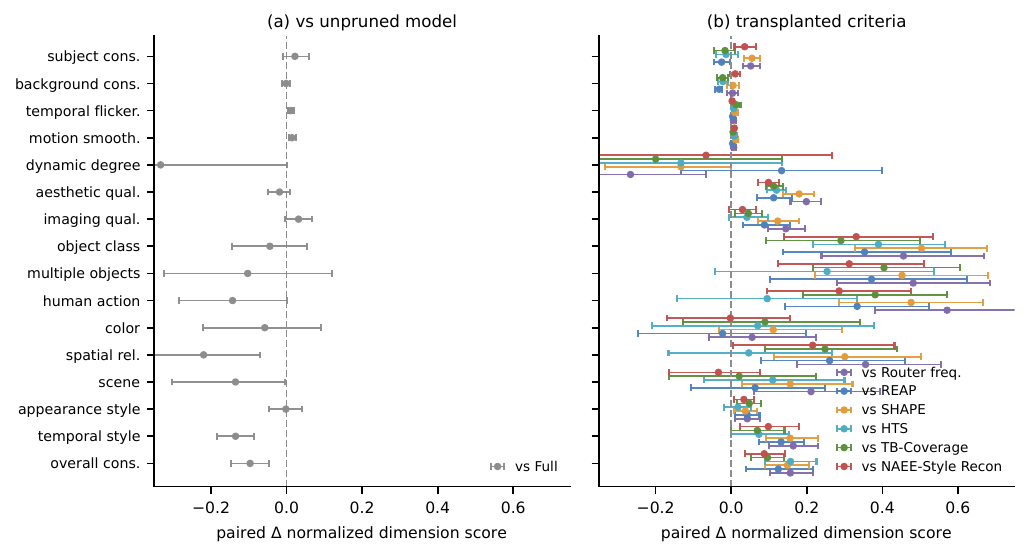}\\[2pt]
\includegraphics[width=0.94\textwidth]{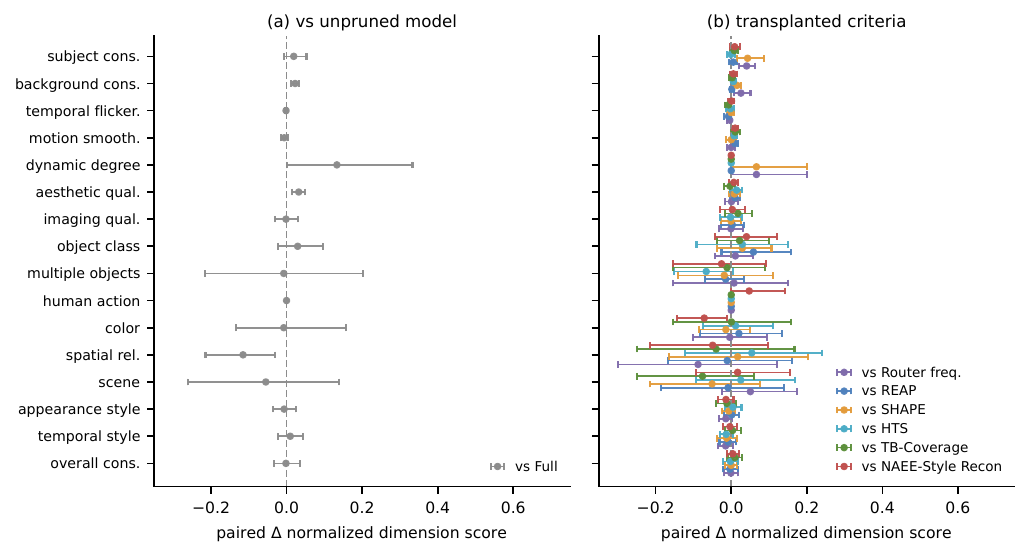}
\vspace{-0.10in}
\caption{Dimension-wise paired differentials at (top) $20\%$ and (bottom) $80\%$ retention, with the same conventions as Figure~\ref{fig:perdim}.}
\vspace{-0.15in}
\label{fig:perdim_extra}
\end{figure}

\begin{figure}[!ht]
\centering
\includegraphics[width=0.47\textwidth]{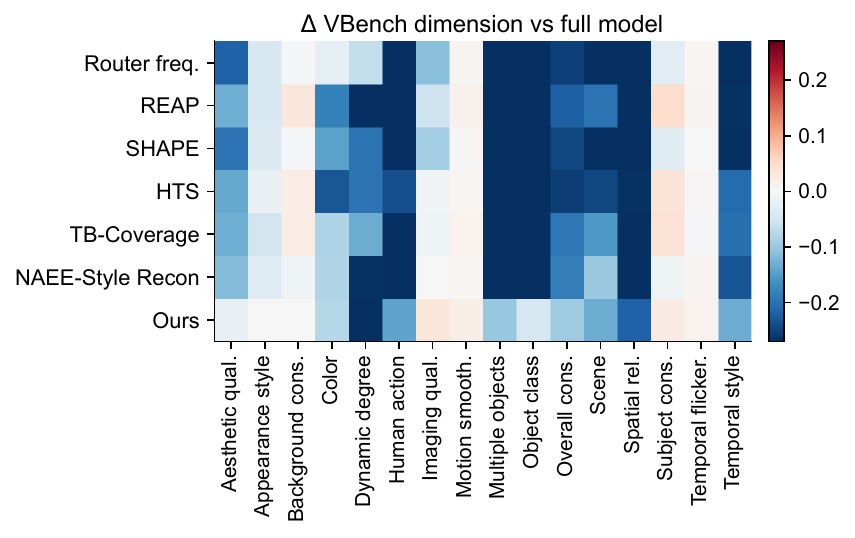}\hfill
\includegraphics[width=0.47\textwidth]{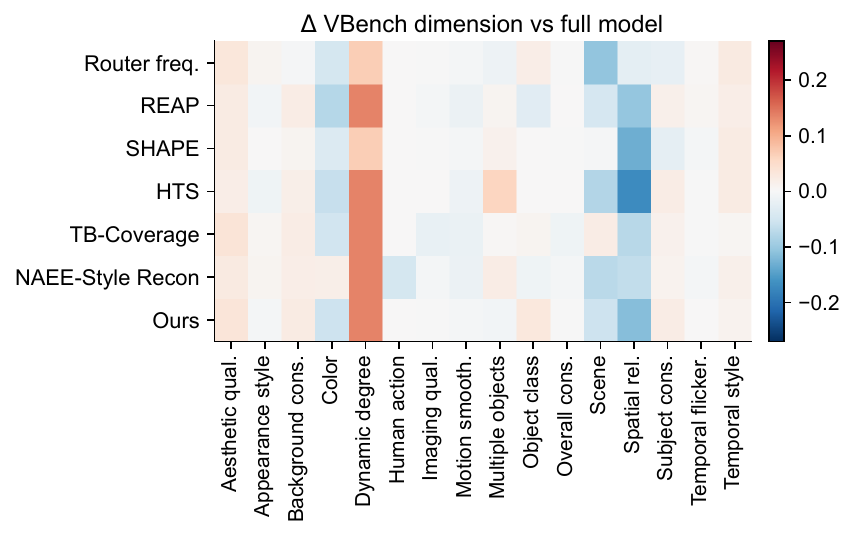}
\vspace{-0.10in}
\caption{Per-dimension deviation from the unpruned baseline at (left) $20\%$ and (right) $80\%$ retention; Figure~\ref{fig:delta} shows the $50\%$ version.}
\vspace{-0.15in}
\label{fig:delta_extra}
\end{figure}

As shown in Figure~\ref{fig:perdim}, differentials relative to the unpruned baseline reflect robust preservation across most dimensions with localized semantic enhancements. Conversely, comparisons against ported LLM baselines demonstrate consistent, statistically significant advantages concentrated in high-level semantic dimensions. The same decomposition at $20\%$ and $80\%$ retention appears in Figures~\ref{fig:perdim_extra} and~\ref{fig:delta_extra}: margins over every ported criterion---including the fidelity route NAEE-Style Recon---persist at $20\%$ and compress toward parity at $80\%$, where only one to three dimensions separate the evaluated criteria.

\section{Geometric Surrogate Formulations and Functional Redundancy}
\label{app:theory}

Overall Diversity Loss operates as a geometric surrogate for functional capacity retention, motivated by empirical deletion responses. Because the signature manifold is high-dimensional and approximately orthogonal, uniform low-rank coverage is unattainable. The ODL metric $1-\cos(\sig_e,\sig_{f(e)})$ directly quantifies the directional discrepancy between excised expert $e$ and its nearest surviving counterpart $f(e)$. Crucially, scale-invariance decouples selection from raw activation magnitude, which our empirical findings demonstrate to be an unreliable proxy for pruning resilience (Section~\ref{sec:ablation}).

\paragraph{Failure modes of norm-based objectives.} 
Consider the global perturbation metric $\mathcal{N}(\mathcal{D}) = \|\sum_{e\in\mathcal{D}}\sig_e\|^2$. In near-orthogonal manifolds, cross-terms vanish ($\langle \sig_i, \sig_j \rangle \approx 0$), reducing the objective to the sum of individual squared norms: $\mathcal{N}(\mathcal{D}) \approx \sum_{e\in\mathcal{D}}\|\sig_e\|^2$. Consequently, norm minimization defaults to excising experts with the smallest individual signature norms. Because signature norm correlates strongly with activation energy ($r=0.91$), this heuristic systematically discards directionally unique experts possessing modest activation scales, precipitating severe degradation across fine-grained semantic dimensions.

\paragraph{Discussion: Anisotropic functional sensitivity.} 
The failure of magnitude-based criteria and the corresponding success of directional coverage motivate an insightful conceptual framing: in video diffusion transformers, representational sensitivity appears anisotropic. High-energy activations predominantly govern broad stylistic and low-frequency texture components that exhibit substantial directional redundancy across the expert bank. Conversely, specialized compositional semantics (e.g., spatial layout and fine-grained object categories) are frequently encoded by directionally unique experts operating at moderate activation scales. Direction-preserving selection via ODL protects these sparse semantic directions, providing a principled foundation for zero-shot expert trimming in video MoEs.

\end{document}